\documentclass[]{style}
\usepackage[most]{tcolorbox}        
\usepackage{amsmath,amssymb}
\usepackage{lipsum}                 
\usepackage{xcolor}                 
\usepackage{fancyhdr}               
\usepackage{booktabs}
\usepackage[table]{xcolor}
\usepackage{url}
\usepackage{graphicx}   
\usepackage{subcaption} 
\usepackage{multirow}
\usepackage{bm}
\definecolor{lightpurple}{RGB}{153, 102, 204}  
\definecolor{lilac}{RGB}{182, 133, 210}        
\usepackage{hyperref}
\hypersetup{
    colorlinks=true,      
    linkcolor=purple,        
    citecolor=lilac,       
    urlcolor=magenta      
}

\definecolor{abstractpurple}{HTML}{9C27B0} 

\renewcommand{\maketitle}{\mymaketitle}
\begingroup
\footnotetext{$*$ Equal contribution, $\dag$ Corresponding author}
\endgroup

\begin{document}

\title{MBench: A Comprehensive Benchmark on Memory Capability for Video World Models}
\author{Shengjun Zhang$^{1,*}$, Zhang Zhang$^{1,2,*}$, Simin Huang$^{1}$, Zhenyu Tang$^{3}$, \\ Hanyang Wang$^{1}$, Chensheng Dai$^{1}$, Min Chen$^{1}$, Yifan Li$^{1}$, Yuxin Li$^{1}$, \\ Yingjie Chen$^{2}$,
Hao Liu$^{2}$, Chen Li$^{2}$, Jing Lyu$^{2}$, Yueqi Duan$^{1,\dag}$}
\affiliation[]{$^{1}$Tsinghua University, $^{2}$WeChat Vision, Tecent Inc., $^{3}$Peking University}

\abstract{
Recent advancements in video-based world models have demonstrated an unprecedented ability to synthesize high-fidelity visual sequences. 
However, a fundamental gap persists between visually plausible video generation and the functional requirements of a world model, particularly in maintaining a stable and reasonable internal state over extended temporal horizons. 
While existing benchmarks primarily emphasize visual quality, motion coherence, and text-video alignment, they largely overlook memory, the core capability of a world model to preserve consistency across long-term horizons and complex interactions.
To address this gap, we present \textbf{MBench}, a comprehensive benchmark dedicated to quantifying and evaluating the memory capability of video world models.  
We systematically decompose the memory capability of video world models into three hierarchical and complementary core dimensions: entity consistency, environment consistency, and causal consistency, which are further refined into 12 quantifiable sub-dimensions for comprehensive characterization of long-term memory. 
Our benchmark is built upon rigorously curated real-captured long videos, and evaluated by rule-based quantitative matrices and VLM to enable objective and comprehensive consistency assessment. 
Extensive evaluations of mainstream state-of-the-art video world models reveal critical systemic limitations of existing methods in long-term state retention, providing a standardized benchmark and clear research direction to advance the field.
}

\checkdata[
\raisebox{-0.2em}{\includegraphics[width=0.025\linewidth]{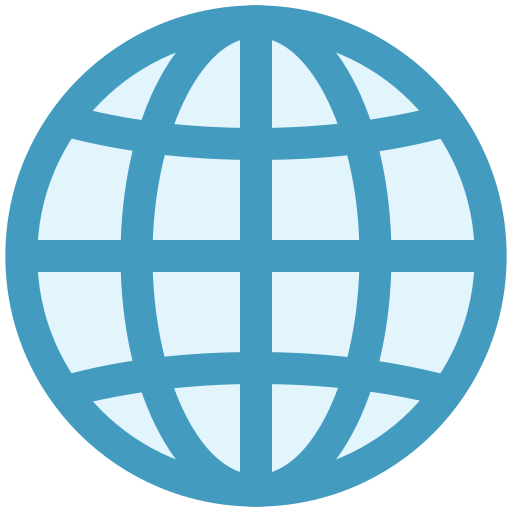}}~~Project Page]{\href{https://peanutup.github.io/MBench-project/}{\texttt{https://peanutup.github.io/MBench-project/}}
\\[-1.5ex]}

\checkdata[
\raisebox{-0.2em}{\includegraphics[width=0.025\linewidth]{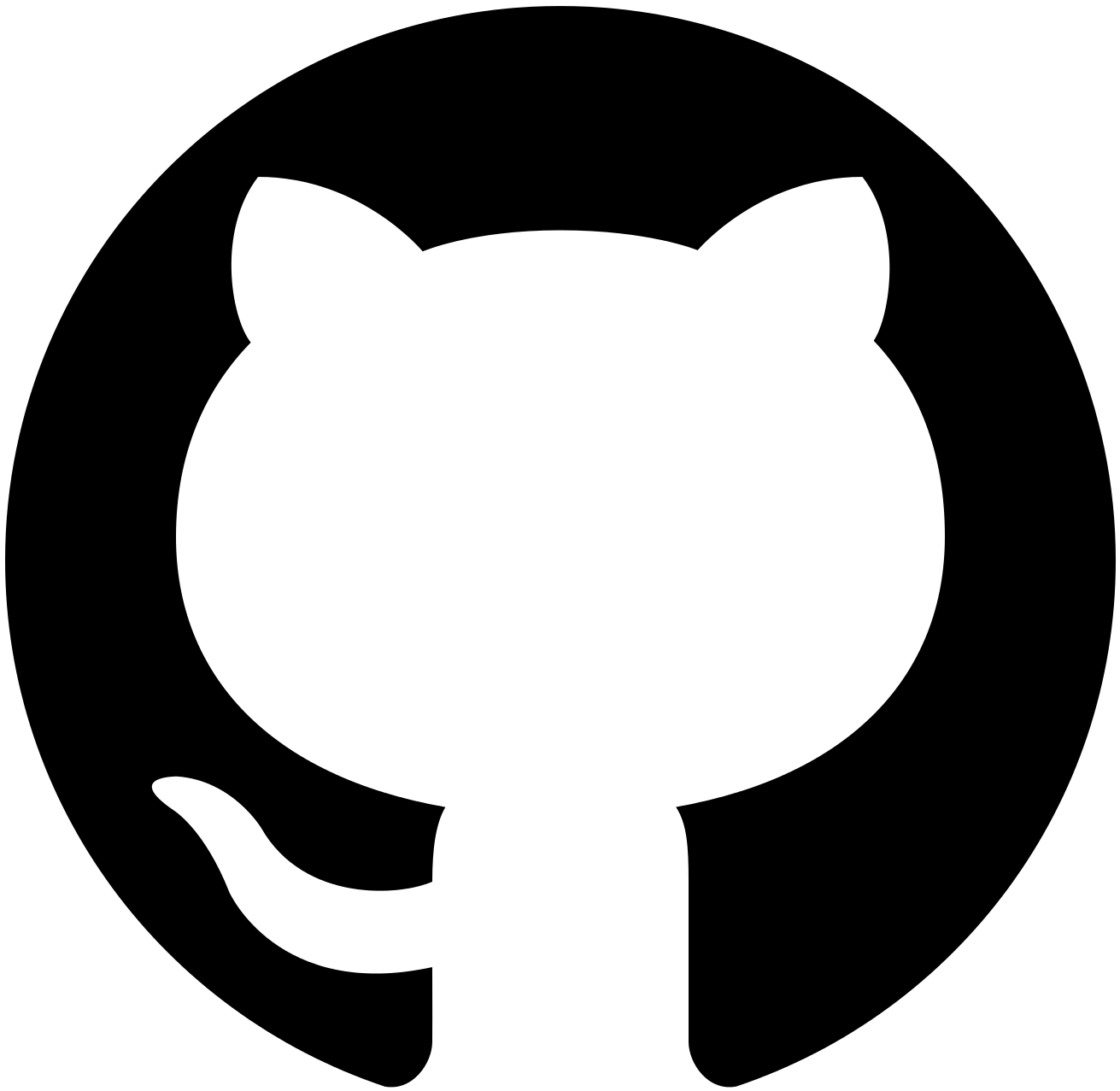}}~~GitHub Repo]{\href{https://github.com/study-overflow/MBench}{\texttt{https://github.com/study-overflow/MBench}}
\\[-1.5ex]}

\checkdata[
\raisebox{-0.2em}{\includegraphics[width=0.025\linewidth]{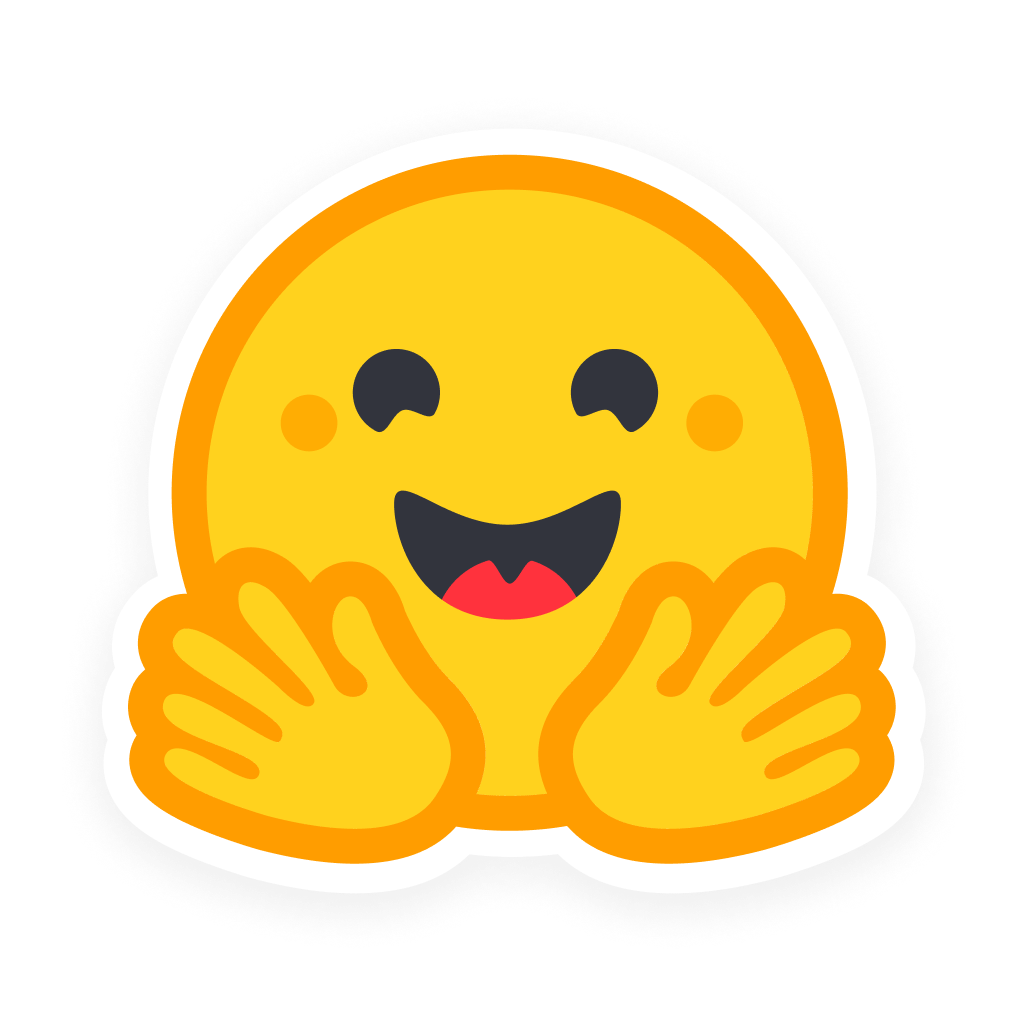}}~~Leaderboard]{\href{https://huggingface.co/spaces/study-overflow/MBench_Leaderboard}{\texttt{https://huggingface.co/spaces/study-overflow/MBench\_Leaderboard}}
\\[-1.7ex]}

\maketitle

\vspace{1em}

\begin{figure}[htbp] 
    \centering
    
    \begin{subfigure}[b]{0.34\textwidth}
        \centering
        \includegraphics[width=\textwidth]{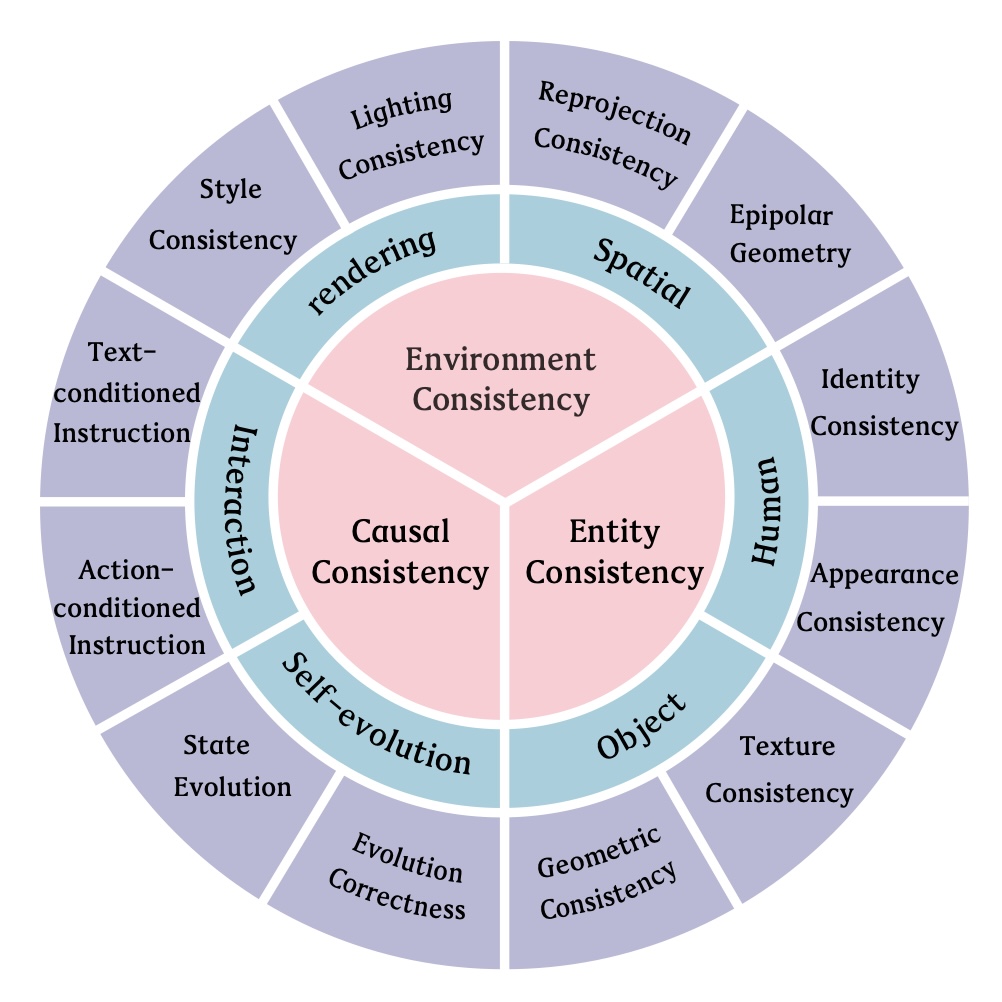} 
        \caption{\textbf{MBench Dimension}}
        \label{fig:MBench Dimension}
    \end{subfigure}
    \hfill 
    \begin{subfigure}[b]{0.30\textwidth}
        \centering
        \includegraphics[width=\textwidth]{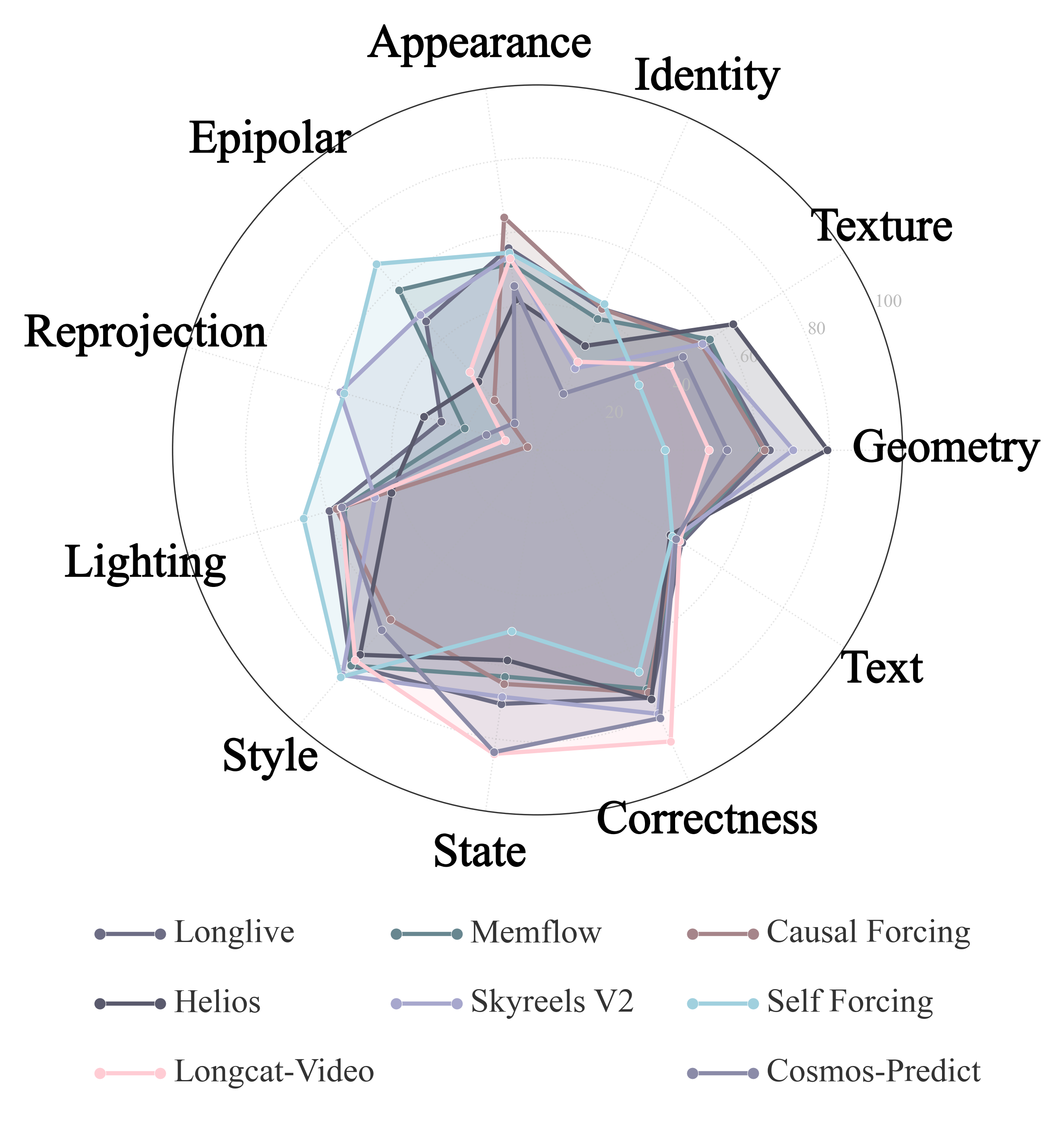}
        \caption{\textbf{Text-Conditioned Models}}
        \label{fig:text-cond comparison}
    \end{subfigure}
    \hfill 
    \begin{subfigure}[b]{0.30\textwidth}
        \centering
        \includegraphics[width=\textwidth]{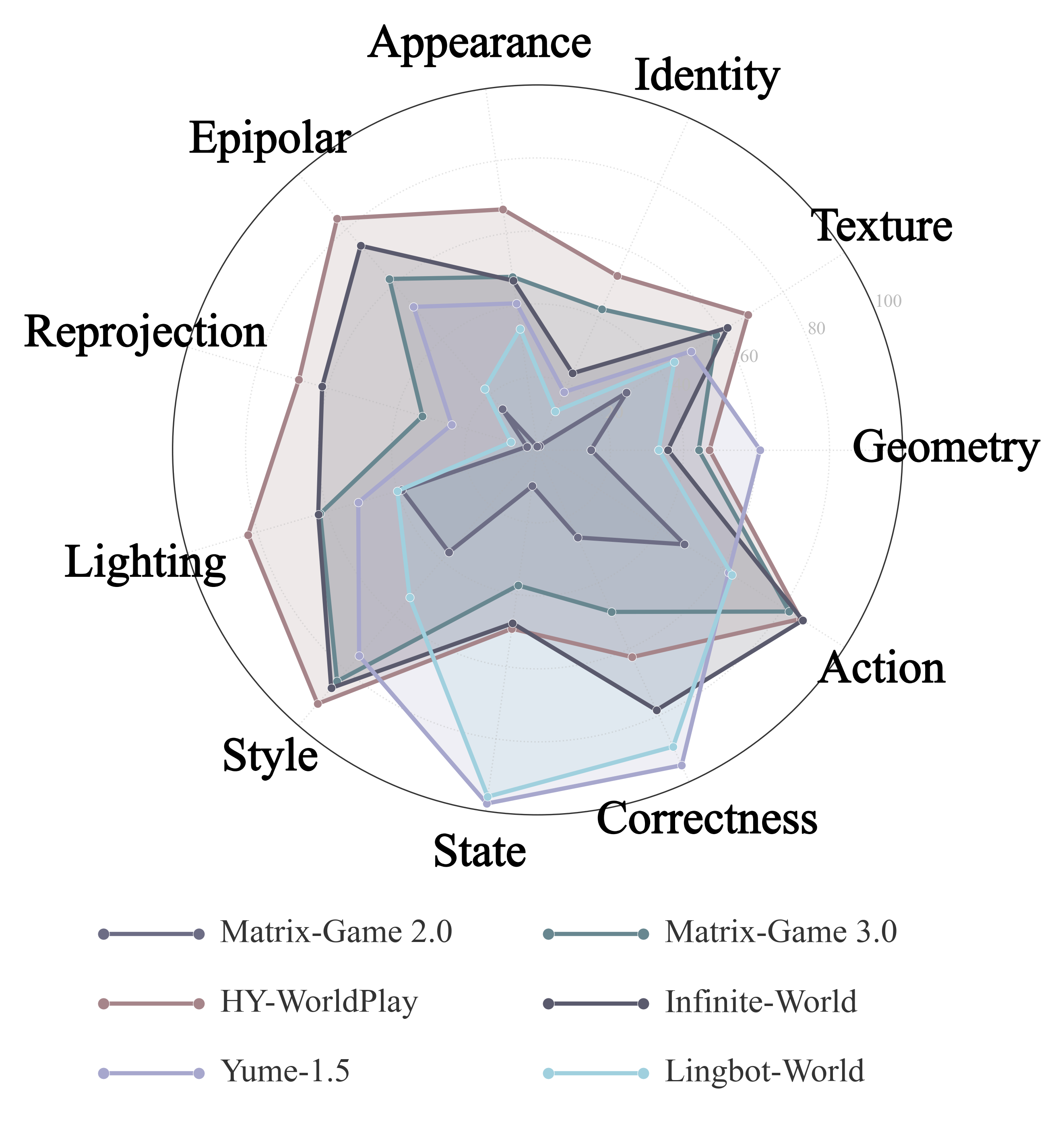}
        \caption{\textbf{Action-Conditioned Models}}
        \label{fig:action-cond comparison}
    \end{subfigure}
    
    \caption{\textbf{Overview of MBench.} (a) MBench consists of a three-level hierarchical taxonomy for comprehensice evaluation. (b) We visualize the evaluation results of 8 video world models with text-conditioned interaction. (c) We visualize the evaluation results of 6 video world models with action-conditioned interaction.}
    \label{fig:three_images}
\end{figure}

\section{Introduction}

\definecolor{checkgreen}{RGB}{0, 150, 0}    
\definecolor{crossred}{RGB}{200, 0, 0}      
\newcommand{\cmark}{\textcolor{checkgreen}{\checkmark}}
\newcommand{\xmark}{\textcolor{crossred}{\times}}

\begin{table}[!t]
\centering
\caption{\textbf{Comparison of Video Generation and World Model Evaluation Benchmarks.}}
\label{tab:benchmark_comparison}
\setlength{\tabcolsep}{5pt} 
\begin{tabular}{lcccccc}
\toprule
\multirow{2}{*}{\textbf{Method}} & \multirow{2}{*}{\textbf{Date}} & \textbf{Entity} & \textbf{Environment} & \textbf{Causal} & \textbf{Action} & \textbf{Long Video} \\
& & \textbf{Consistency} & \textbf{Consistency} & \textbf{Consistency} & \textbf{Condition} & \textbf{Support}\\
\midrule
VBench~\cite{huang2023vbench}          & 2023.11 & $\cmark$ & $\cmark$ & $\xmark$ & $\xmark$ & $\xmark$ \\
ChronoMagicBench & 2024.06 & $\xmark$ & $\xmark$ & $\xmark$ & $\xmark$ & $\xmark$ \\
T2V--CompBench  & 2024.07 & $\cmark$ & $\xmark$ & $\xmark$ & $\xmark$ & $\xmark$ \\
VBench++~\cite{huang2025vbench++}        & 2024.11 & $\cmark$ & $\cmark$ & $\xmark$ & $\xmark$ & $\xmark$ \\
OpenS2V--Eval   & 2025.06 & $\cmark$ & $\xmark$ & $\xmark$ & $\xmark$ & $\xmark$ \\
VBench 2~\cite{zheng2025vbench2}        & 2025.08 & $\cmark$ & $\cmark$ & $\xmark$ & $\xmark$ & $\xmark$ \\
LoCoT2V--Bench  & 2026.01 & $\xmark$ & $\xmark$ & $\xmark$ & $\xmark$ & $\cmark$ \\
\midrule
WorldModelBench~\cite{Li2025WorldModelBench} & 2025.02 & $\xmark$ & $\xmark$ & $\xmark$ & $\xmark$ & $\xmark$  \\
WorldScore~\cite{duan2025worldscore}      & 2025.11 & $\xmark$ & $\cmark$ & $\xmark$ & $\xmark$ & $\xmark$ \\
WorldBench~\cite{upadhyay2025worldbench}      & 2026.01 & $\xmark$ & $\xmark$ & $\xmark$ & $\xmark$ & $\xmark$  \\
MIND~\cite{ye2026mind}            & 2026.02 & $\xmark$ & $\cmark$ & $\xmark$ & $\cmark$ & $\xmark$ \\
STEVO-Bench~\cite{stevo-bench2026}    & 2026.03 & $\xmark$ & $\xmark$ & $\cmark$ & $\xmark$ & $\xmark$ \\
Omni-WorldBench~\cite{wu2026omni-worldbench} & 2026.03 & $\xmark$ & $\cmark$ & $\xmark$ & $\cmark$ & $\xmark$ \\
WorldMark~\cite{xu2026worldmark} & 2026.04 & $\xmark$ & $\cmark$ & $\xmark$ & $\cmark$ & $\cmark$ \\
\midrule
\rowcolor{gray!10} 
\textbf{MBench (Ours)} & \textbf{2026.05} & $\cmark$ & $\cmark$ & $\cmark$ & $\cmark$ & $\cmark$ \\
\bottomrule
\end{tabular}

\end{table}

\begin{figure}[t]
    \centering

    \begin{subfigure}{0.9\linewidth}
        \centering
        \includegraphics[width=\linewidth]{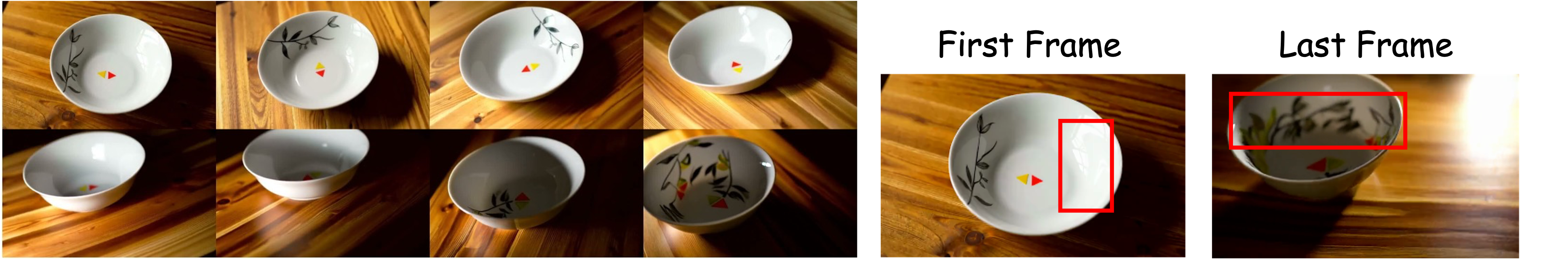}
        \caption{\textbf{Entity Consistency}}
        \label{fig:entity case}
    \end{subfigure}

    \vspace{0.5em}
    
    \begin{subfigure}{0.9\linewidth}
        \centering
        \includegraphics[width=\linewidth]{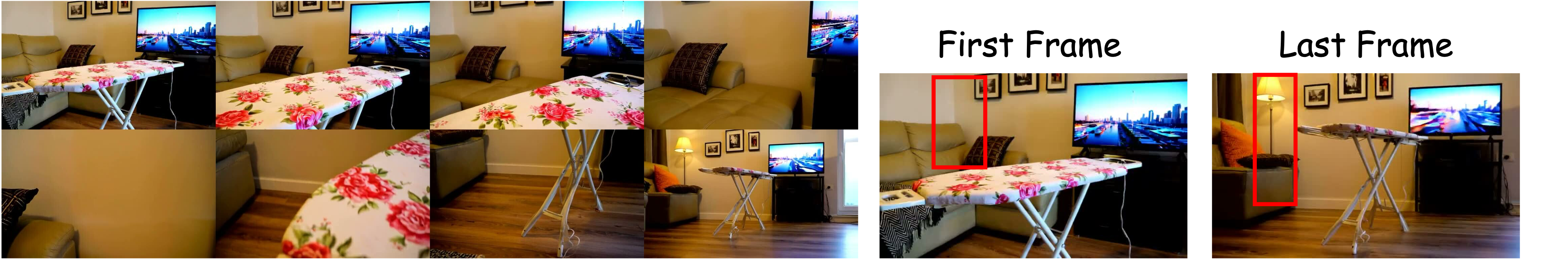}
        \caption{\textbf{Environment Consistency}}
        \label{fig:environment case}
    \end{subfigure}
    
    \vspace{0.5em}
    
    \begin{subfigure}{0.9\linewidth}
        \centering
        \includegraphics[width=\linewidth]{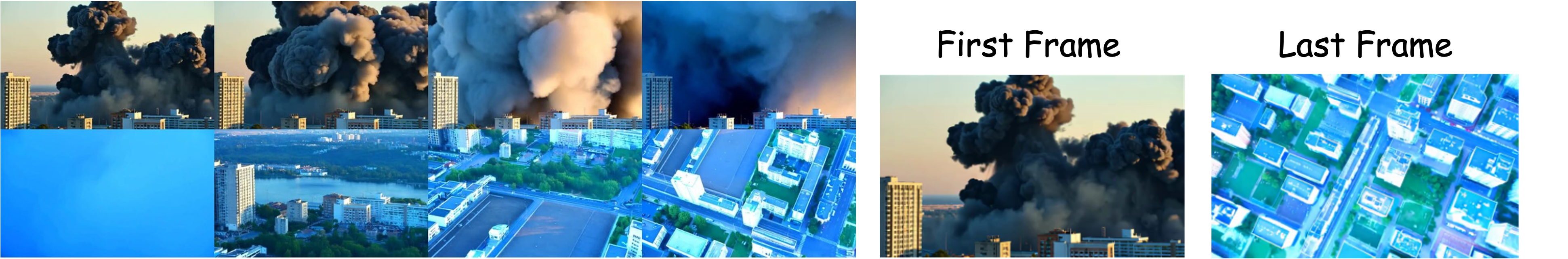}
        \caption{\textbf{Causal Consistency}}
        \label{fig:causal case}
    \end{subfigure}

    \caption{\textbf{Cases of different evaluation dimensions.}}
    \label{fig:memory cases}
\end{figure}

In recent years, video generation models~\cite{openai2025sora2, cogvideox2025, seedance2026, wan2025, google2025veo3} have achieved remarkable breakthroughs in synthesizing high-fidelity and temporally coherent realistic videos, establishing a solid foundation for the development of powerful video-based world models~\cite{bruce2024genie, hyworld2025, mao2025yume}. 
These models have accelerated advancements across diverse high-impact domains, including robotics~\cite{genieenvisioner2025, unitree2025unifolmwma0, cosmos-predict2025}, autonomous driving~\cite{GAIA-1, DriveDreamer2023}, and interactive immersive gaming~\cite{matrix-game-3.0, yan2025}. 
Rather than merely producing visually plausible standalone video clips, a functional world model must faithfully maintain the inherent dynamics, physical laws, and semantic rules of the real world, and reliably support long-horizon prediction, reasoning, and interaction. 
While recent video generation models can produce minute-long visual sequences~\cite{huang2025selfforcing, MAGI-1, longcat-video, yang2025longlive}, a fundamental gap persists between visual synthesis and world modeling, which requires long-term memory to maintain stable, consistent internal representations of world states across extended temporal horizons and interaction.

This critical gap is further exacerbated by the inadequacy of existing evaluation benchmarks for video world models. 
Currently, dominant evaluation protocols~\cite{huang2023vbench, zheng2025vbench2, duan2025worldscore, upadhyay2025worldbench} predominantly center on visual quality, motion coherence, and text-video alignment. 
While these metrics effectively quantify how realistic or well-aligned a generated video appears, they entirely overlook the long-term memory of world models. 
Unlike one-off short video generation, a functional world model must reliably retain the persistent properties of the world, where the identity and attributes of entities should not arbitrarily change, the layout and structure of the environment should remain stable across camera motions and scene dynamics, and the causal consequences of prior events should consistently govern future scene evolution. 
However, existing benchmarks fail to systematically challenge or measure these core memory capabilities. 
As shown in Table~\ref{tab:benchmark_comparison}, most benchmarks are limited to short-horizon generation tasks, and lack structured decomposition of memory-related consistency requirements. 
As a result, the field currently lacks a standardized way to distinguish between models that merely generate visually appealing frames, and those that can truly simulate a predictable real world.

To this end, we present MBench, a comprehensive benchmark dedicated to systematically quantifying and evaluating the long-term memory capability of generative video world models. 
As illustrated in Figure~\ref{fig:MBench Dimension}, our benchmark is built upon a three-level hierarchical taxonomy that decomposes world model memory into three foundational, complementary, and functionally critical dimensions: 
\begin{enumerate}
    \item Entity Consistency focuses on the model’s ability to retain the persistent identity and attributes of individual participants throughout the generated sequence, which is bifurcated into Object Consistency and Human Consistency, ensuring that both inanimate items and human agents maintain their unique characteristics over time. As shown in Figure~\ref{fig:entity case}, a model with robust entity memory must preserve the specific texture and geometric integrity of an object (e.g., the intricate patterns on a bowl) even during complex camera movements or occlusions.
    \item Environment Consistency measures the stability of the ``stage'' upon which the simulation unfolds. It consists of two primary second-level modules: Spatial Consistency, which evaluates the stability of 3D scene layouts and relative positioning, and Rendering Consistency, which monitors global scene properties like lighting and stylistic coherence. A typical failure in this dimension is illustrated in the center of Figure~\ref{fig:environment case}, where the spatial arrangement of furniture or background elements may drift or collapse as the perspective shifts.
    \item Causal Consistency assesses whether the model can uphold the logical relationships between prior states and future dynamics, ensuring the world follows established physical and semantic rules. This dimension is divided into Self-evolution, which tests the model’s internal adherence to inherent physical laws, and Interaction, which measures the model’s ability to remember and propagate the consequences of external prompts or actions. A typical failure in this dimension is illustrated in Figure~\ref{fig:causal case}, where after an explosion, the camera passes through the resulting cloud of dust and smoke, only to reveal a perfectly intact street lined with orderly houses instead of the ruins and debris that the explosion should have caused. This exposes a fundamental breakdown in causal reasoning: the model fails to carry forward the logical consequences of a destructive event into the subsequent visual state.
\end{enumerate}
This systematic taxonomy enables a multi-faceted evaluation that spans from individual object-level details to holistic scene-level layouts, while simultaneously encompassing both the persistence of static properties and the logical validity of dynamic temporal evolutions.
Together, this hierarchical framework enables MBench to provide a granular diagnostic of a world model's memory, moving beyond simple visual quality to probe the underlying logical and structural stability of the simulated world.


Our benchmark is built upon a rigorously curated dataset of real-captured long videos, filtered to select clips that pose meaningful, graded challenges to long-term temporal stability, where we strategically introduce contextual cues designed to provoke and expose latent memory failures during the generative process. 
To ensure a granular assessment, we employ a hybrid evaluation strategy that integrates normalized heuristic metrics with VLM-augmented Visual Question Answering (VQA).
Specifically, MBench comprises over 1040 standardized evaluation cases, through which we have conducted an extensive audit of 12 state-of-the-art video world models. 
As illustrated in Figure~\ref{fig:text-cond comparison} and~\ref{fig:action-cond comparison}, our experimental results uncover prevalent systemic vulnerabilities in existing methods.
Most advanced models exhibit significant temporal drift and forgetfulness as the simulation horizon extends. 
By providing a rigorous, multi-faceted diagnostic tool, MBench reveals that current generative paradigms struggle to maintain a persistent world state under interaction or long-range evolution. Ultimately, our work establishes a standardized evaluation protocol and delineates a clear research trajectory for the development of truly consistent and reliable video world models.

\section{Related Works}

\subsection{Video Generation and World Models}
\textbf{Video Generation Models.}
Early video generation built on GANs~\cite{vondrick2016generating, tulyakov2018mocogan, clark2019adversarial} and VAE-based autoregressive transformers~\cite{yan2021videogpt, villegas2022phenaki, yu2023magvit}, producing short, low-resolution clips. Video Diffusion Models~\cite{ho2022video} extended image diffusion to the temporal axis, and subsequent works scaled this paradigm through cascaded super-resolution~\cite{ho2022imagen, singer2022make}, latent diffusion~\cite{blattmann2023stable, guo2023animatediff, chen2023videocrafter1}, and the diffusion transformer~\cite{peebles2023scalable}. Sora~\cite{liu2024sora} pioneered large-scale DiT training and demonstrated minute-level photorealistic synthesis, catalyzing a wave of foundation video models including CogVideoX~\cite{cogvideox2025}, HunyuanVideo~\cite{kong2024hunyuanvideo}, Wan~\cite{wan2025}, Veo-3~\cite{google2025veo3}, Seedance 2.0~\cite{seedance2026}, and MAGI-1~\cite{MAGI-1}. To break the short-clip ceiling, recent work couples diffusion with autoregressive rollout for streaming generation, exemplified by Self Forcing~\cite{huang2025selfforcing}, LongLive~\cite{yang2025longlive}, LongCat-Video~\cite{longcat-video}, and SkyReels-V2~\cite{chen2025skyreelsv2}. Despite this progress, mainstream video generators are optimized for perceptual quality and prompt alignment, and produce standalone clips rather than coherent world dynamics.

\textbf{World Models.}
A world model learns environment dynamics and predicts future states under given conditions, originating in reinforcement learning~\cite{ha2018recurrent} and extended through recurrent~\cite{hafner2023mastering}, transformer~\cite{micheli2022transformers}, and diffusion-based~\cite{alonso2024diffusion} dynamics. Large-scale video pretraining has since lifted this paradigm to open-domain settings, yielding three application-driven lines: interactive game-style worlds with discrete actions~\cite{bruce2024genie, matrix-game-2, matrix-game-3, hyworld2025, yume-1.5, yan2025, helios, lingbot-world, wu2026infiniteworld}, driving simulators conditioned on ego-trajectories~\cite{GAIA-1, DriveDreamer2023, gao2024vista, cosmos-predict2025}, and robotic foundation models conditioned on manipulation policies~\cite{yang2023learning, genieenvisioner2025, unitree2025unifolmwma0}. Across all three, the bottleneck is no longer per-frame fidelity but long-horizon state consistency: as rollouts extend, models forget previously generated content and drift away from grounded dynamics. This bottleneck motivates the systematic study of memory in video world models.

\subsection{Memory Modeling in Video World Models}

Long-horizon consistency is the defining challenge of video world models. Early long-video synthesis approached this through causal masking~\cite{villegas2022phenaki}, hierarchical diffusion~\cite{yin2023nuwa}, and temporal co-denoising~\cite{wang2023gen}, but treated all history uniformly and left consistency to emerge implicitly from data. Recent approaches make memory a first-class design target and can be organized by where world-state information is stored.

\textbf{In-context memory.}
In-context memory models retain history as attention activations within an extended or restructured context window. Compression-based designs allocate token budget asymmetrically across time, encoding recent frames densely while summarizing distant ones~\cite{gu2025long, zhang2025packing, jin2024pyramidal}. Streaming designs instead extend rollout horizons via sliding-window autoregression~\cite{chen2025skyreelsv2, longcat-video, henschel2025streamingt2v, qiu2023freenoise, kim2024fifo}.

\textbf{External memory.} 
External memory architectures offload history into an explicit bank that is queried on demand, with retrieval keys ranging from camera pose~\cite{xiao2025worldmem} to semantic relevance~\cite{ji2025memflow, chen2025learning} and pose-free hierarchical descriptors~\cite{matrix-game-3, wu2026infiniteworld, zhu2025memorize}. This architecture decouples context length from compute and is naturally suited to scenarios where the model must revisit previously generated regions.

\textbf{Parametric memory.} 
Parametric memory approaches absorb history directly into model weights at inference time. Test-time training video models~\cite{dalal2025one} treat long-context tokens as an online dataset and adapt internal MLPs accordingly. A separate line of work tackles autoregressive train--test mismatch through auxiliary training schemes that expose models to noised or self-generated history~\cite{chen2024diffusion, song2025history, huang2025selfforcing, yang2025longlive, zhu2026causal-forcing}, which improve robustness to imperfect history but are orthogonal to the memory architecture itself.

\subsection{Evaluation for Video Generation and World Models}

Early video generation evaluation relied on distribution-level metrics such as Inception Score~\cite{salimans2016improved}, Fréchet Inception Distance~\cite{heusel2017gans}, Fréchet Video Distance~\cite{unterthiner2018towards}, LPIPS~\cite{zhang2018unreasonable}, and CLIPSIM~\cite{wu2021godiva}, which capture frame realism and text--video alignment but reveal little about temporal or semantic consistency. Comprehensive benchmarks decompose quality into fine-grained perceptual dimensions~\cite{huang2023vbench, huang2025vbench++, zheng2025vbench2, liu2024evalcrafter, liu2023fetv, sun2025t2v, han2025video}, with specialized variants probing physics~\cite{bansal2024videophy, meng2024towards, guo2025t2vphysbench}, time-lapse coherence~\cite{yuan2024chronomagic}, and long-form generation~\cite{ji2024t2vbench, zheng2025locot2v}. These protocols target content-creation quality rather than persistent world-state retention.

A recent wave of benchmarks targets world models directly, evaluating physical fidelity and instruction following~\cite{Li2025WorldModelBench, duan2025worldscore, upadhyay2025worldbench}, with two concurrent efforts probing memory under specific conditions~\cite{ye2026mind, stevo-bench2026}. As summarized in Table~\ref{tab:benchmark_comparison}, however, no existing benchmark jointly covers entity, environment, and causal consistency, supports both text- and action-conditioning, and probes long videos with explicit memory triggers. MBench fills this gap with a three-axis memory taxonomy and Trigger-Conditioned Scoring, which prevents conservative models from inflating scores by avoiding the challenges the benchmark is designed to expose.

\section{Benchmark Construction}

In this section, we present the detailed design of MBench, a comprehensive benchmark dedicated to evaluating the long-term memory capability of video world models. 


\begin{figure}[t]
    \centering
    \begin{subfigure}{0.48\linewidth}
        \centering
        \includegraphics[width=\linewidth]{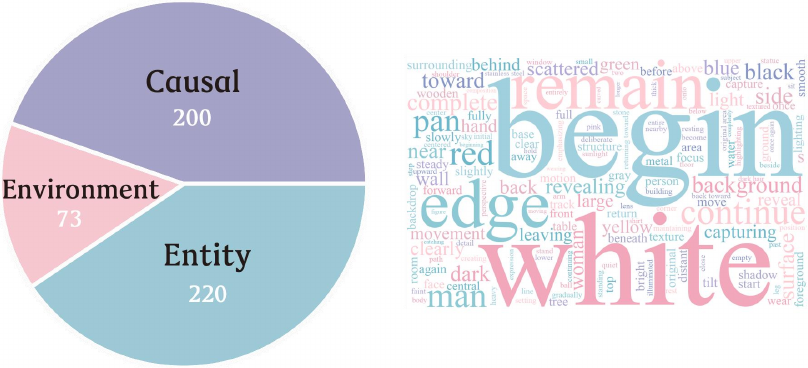}
        \caption{\textbf{Prompts for Action-Conditioned Models}}
    \end{subfigure}
    \hfill
    \begin{subfigure}{0.48\linewidth}
        \centering
        \includegraphics[width=\linewidth]{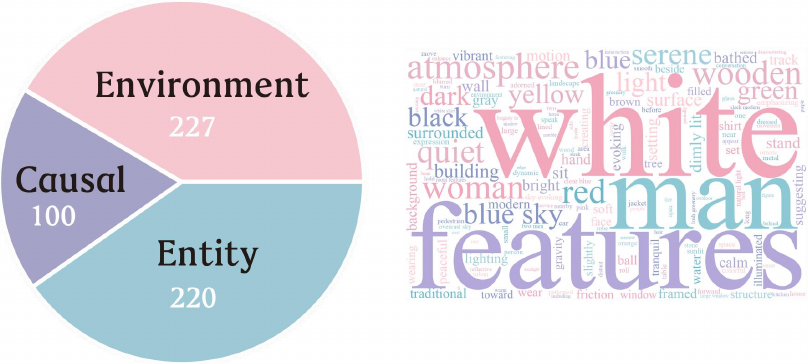}
        \caption{\textbf{Prompts for Text-Conditioned Models}}
    \end{subfigure}
    \caption{\textbf{Prompt Statistics.}}
\end{figure}

%
%
%

\subsection{Data collection}
We aggregate data from five diverse real-world video datasets: DL3DV~\cite{ling2024dl3dv}, Tanks and Temples~\cite{tanks-temples2017}, OpenHumanVID~\cite{li2024openhumanvid}, SpatialVID~\cite{wang2025spatialvid}, and Physics-aware-video~\cite{physics-aware-video}. These datasets collectively cover a wide range of scenarios including indoor and outdoor environments, human-object interactions, dynamic camera motions, and physical state transitions. 
Video durations range from 5 seconds to 15 minutes, providing sufficient temporal context for evaluating long-term memory retention.

Furthermore, we use a Vision Language Model (VLM) to systematically select videos that pose meaningful challenges to the three dimensions of memory consistency. 
The model is specifically prompted to evaluate the video's complexity across three core challenges: entity consistency, by identifying entities undergoing occlusion, attribute changes, or complex motion trajectories; environment consistency, by detecting significant camera motions, scene layout changes, or dynamic background elements; and causal consistency, by recognizing physical state transitions, or causal event chains.

\subsection{Prompt Generation}

For each curated video clip, we generate a detailed description of the overall scene and main activity, which covers precise attributes, positions of entities, and chronological sequence of key events.

\subsubsection{Multi-Granularity Caption Generation}
 For video continuation models, we further segment the complete structured description into five semantically coherent consecutive paragraphs along the temporal axis, and prepend designed camera control instructions to each paragraph. 
These instructions introduce memory-triggering camera motions including translation, rotation, zooming and dynamic occlusion, which force the model to maintain consistent world state representations under changing viewing conditions, thus effectively measuring its long-term memory capability. 
These multi-granularity captions also serve as the foundation for generating comprehensive ground truth question-answering pairs, ensuring our assessment covers all aspects of memory consistency.

\subsubsection{Action Sequence Generation}

For action-conditioned world models, we adopt an Exit-and-Reenter paradigm, which is specifically designed to rigorously test the model's ability to retain the complete state of entities when they are temporarily outside the camera's field of view. 
For a video with initial frame $I_0$ and a predefined target entity $E$, we construct a standardized action sequence $A = [a_1, a_2, \dots, a_T]$ consisting of three consecutive phases: first, the exit phase, where the camera moves along a predefined trajectory to make the target entity $E$ completely exit the field of view; second, the wait phase, where the camera remains stationary for a fixed duration while $E$ stays entirely invisible; and third, the reenter phase, where the camera moves along the exact reverse trajectory to bring $E$ back into the field of view and return to the initial camera position and orientation.

\section{Benchmark Evaluation}\label{sec:mbench-evaluation}

\subsection{Automated Evaluation Pipeline}
We design a multi-dimensional, fully automated evaluation pipeline to quantify the memory capability of video world models across the three core dimensions: entity consistency, environment consistency, and causal consistency. 
For each dimension, we decompose it into fine-grained sub-dimensions and adopt specialized quantitative metrics tailored to different evaluation objectives. 
This hierarchical evaluation framework enables us to provide a comprehensive and interpretable assessment of model performance, identifying specific strengths and weaknesses of different approaches.

A critical challenge in evaluating generative world models is accounting for generative stochasticity: different models may exhibit varying levels of responsiveness to memory triggers. Some models may generate static or overly conservative content that avoids memory challenges entirely, leading to inflated consistency scores that do not reflect true memory capability. 
To address this fundamental issue, we propose a Trigger-Conditioned Scoring  mechanism that decouples performance into two complementary metrics and computes their harmonic mean as the final ranking score.
Our evaluation proceeds in two sequential stages:
\begin{enumerate}
    \item Trigger Verification: For each test sample, we first use a VLM to verify whether the model has successfully executed the memory-triggering event specified in the prompt, such as object occlusion and camera motion causing entity exit. We record a binary trigger label for each sample.
    \item Consistency Evaluation: Only for samples that have successfully triggered the memory challenge, we compute the correspondingthe evaluation matrix described in the following subsections. For samples that fail to trigger the challenge, no consistency score is calculated, as they do not provide valid information about the model's memory capability.
\end{enumerate}
Based on this two-stage process, we define two core metrics.
First, the trigger coverage matrix $C^{\text{trig}}$ measures the model's ability to follow instructions and simulate complex dynamic scenarios:
\begin{equation}
C^{\text{trig}}_k = \frac{N^{\text{trig}}_k}{N_{\text{total}}}
\end{equation}
where $N_{\text{total}}$ is the total number of test samples in the benchmark, and $N^{\text{trig}}_k$ is the total number of triggered samples of the $k$-th evaluation score for the model.
Second, the memory reliability matrix $S^{\text{rel}}$ measures the average consistency performance of a model exclusively on samples where it successfully engages with the memory challenge:
\begin{equation}
S^{\text{rel}}_k = \frac{1}{N^{\text{trig}}_k} \sum_{i=1}^{N^{\text{trig}}_k} S_k(i)
\end{equation}
where $S_k(i)$ is the $k$-th evaluation score for sample $i$.
To prevent models from achieving high rankings by simply avoiding memory challenges, we compute the final M-Score as the harmonic mean of $S^{\text{rel}}_k$ and $C^{\text{trig}}_k$, which effectively balances memory reliability and instruction responsiveness:
\begin{equation}
\text{M-Score}_k = 2 \times \frac{S^{\text{rel}}_k \times C^{\text{trig}}_k}{S^{\text{rel}}_k + C^{\text{trig}}_k}
\end{equation}
This scoring mechanism penalizes overly conservative models with low trigger coverage while rewarding models that maintain high consistency under frequent, complex world-state transitions. 
In our experiments, we report M-Score of each dimension for a comprehensive analysis of model performance.

\subsection{Entity Consistency}
Entity consistency evaluates the model's ability to maintain the persistent identity, attributes, and spatial-temporal trajectories of individual entities throughout the generation process. We distinguish between general object consistency and human-specific consistency, as human entities require additional evaluation of identity preservation beyond basic visual attributes.

\subsubsection{Object Consistency}
Object consistency measures the stability of geometric and textural attributes of non-human entities across the video sequence.





\textbf{Geometry Consistency.}
To rigorously evaluate the long-term geometric self-consistency of generated videos, particularly in scenarios where objects exit and subsequently re-enter the camera's field of view (FoV), we introduce the Geometry Consistency Score. 
This evaluation specifically targets generated sequences that feature \textit{departure-return} camera trajectories. 
Within such a continuous trajectory, we establish fold pairs by identifying the most viewpoint-aligned corresponding frames. 
Let frame $i$ denote a frame from the forward pass (\textit{departure}) and frame $j$ denote its paired counterpart from the return pass (\textit{return}) exhibiting the closest viewpoint proximity.
For each target object present in a fold pair, we extract its precise spatial mask $M_i$ in the reference frame $i$ utilizing SAM 2~\cite{ravi2024sam2}. 
To quantify the geometric preservation of the object over the long temporal span, we project the return frame $I_j$ to the viewpoint of the forward frame $I_i$ via a warping operation, denoted as $\mathcal{W}_{j \to i}(\cdot)$. 
The Geometry Consistency Score is then formulated by calculating the SSIM strictly within the object's masked region. Mathematically, it is defined as:
\begin{equation}
    \text{Geometry Consistency Score} = \text{SSIM}\Big(I_i \odot M_i, \, \mathcal{W}_{j \to i}(I_j) \odot M_i\Big)
\end{equation}
where $I_i$ and $I_j$ denote the visual tensors of frame $i$ and frame $j$ respectively, and $\odot$ represents the Hadamard product. 
By constraining the SSIM computation to the intersection of the warped frame and the SAM 2~\cite{ravi2024sam2} mask $M_i$, this metric effectively isolates the target object from the background. 
Consequently, it provides a robust quantitative measure of whether the video generation model can faithfully preserve the intrinsic structural and geometric details of an object after undergoing complex, long-duration occlusions or boundary truncations.

\textbf{Texture Consistency.}
Texture Consistency assesses whether the visual attributes of a specific object—including its color, texture, and intricate patterns—remain invariant throughout generation. 
During evaluation, we utilize SAM 2~\cite{ravi2024sam2} to propagate precise segmentation masks across the entire sequence based on the provided anchor. 
We then extract semantic feature vectors $\{f_i\}$ from the masked regions using the DINOv2 backbone~\cite{oquab2023dinov2}. Following the same \textit{departure-return} paradigm as our human consistency metrics, we move beyond simple adjacent-frame comparisons and calculate the score as the average cosine similarity between each frame's feature and the global track centroid:
\begin{equation}
\text{Texture Consistency Score} = \frac{1}{N} \sum_{i=1}^{N} \frac{f_{i} \cdot \bar{f}}{\|f_{i}\| \|\bar{f}\|}
\end{equation}
where $f_{i}$ denotes the DINOv2 feature of the target object at frame $i$, and $\bar{f} = \frac{1}{N} \sum_{j=1}^{N} f_{j}$ represents the global mean embedding of the object throughout the video. This formulation effectively quantifies the model's ability to "remember" and maintain the fine-grained visual identity of an object across long-term temporal horizons and complex spatial transitions.

\subsubsection{Human Consistency}
Human consistency extends object consistency with additional evaluation of identity preservation, which is critical for scenarios involving human characters.


\textbf{Identity Consisitency.}
Identity Consistency evaluates whether the identity of a human subject remains stable and recognizable throughout the generated sequence, particularly across temporal gaps or occlusions. To quantify this, we implement a multi-object tracking pipeline utilizing a pre-trained ArcFace model~\cite{deng2019arcface}. 
Specifically, we uniformly sample 32 frames from the video and detect all human faces in each frame. For identity maintenance, we maintain a feature track $\mathcal{T}_k$ for each detected subject $k$, which stores the history of facial embeddings $\{\phi_{k,i}\}$. During the tracking process, we employ a rolling-average matching strategy: each new detection is compared against both the subject's most recent embedding and its cumulative rolling average embedding using cosine similarity. We then utilize greedy bipartite matching with a threshold of 0.15 to assign detections to existing tracks or initiate new ones. 
Unlike simple adjacent-frame comparisons which are prone to gradual identity drift, we define the Identity Consistency Score as the average cosine similarity between all embeddings within a track and their respective global centroid (mean embedding), averaged across all $K$ subjects:
\begin{equation}
\text{Identity Consistency Score} = \frac{1}{K} \sum_{k=1}^{K} \left( \frac{1}{N_k} \sum_{i=1}^{N_k} \frac{\phi_{k,i} \cdot \bar{\phi}_k}{\|\phi_{k,i}\| \|\bar{\phi}_k\|} \right)
\end{equation}
where $K$ is the total number of unique human tracks, $N_k$ is the number of detections in track $k$, $\phi_{k,i}$ denotes the $i$-th facial embedding in track $k$, and $\bar{\phi}_k = \frac{1}{N_k} \sum_{j=1}^{N_k} \phi_{k,j}$ represents the centroid embedding for that subject. This centroid-based metric effectively captures the long-term stability of human identities and their robustness to \textit{departure-return} scenarios.


\textbf{Appearance Consistency.}
Appearance Consistency evaluates the temporal stability of a subject's holistic visual attributes, such as clothing, hairstyle, and accessories. Building on the multi-object tracks established in the Identity Consistency phase, we employ SAM 2~\cite{ravi2024sam2} to extract precise full-body masks for each identified subject. We then utilize the DINOv2 backbone~\cite{oquab2023dinov2} to generate semantic embeddings from these masked regions. Following the same \textit{departure-return} paradigm and centroid-based similarity formulation as defined in the Identity Consistency section, this metric quantifies the invariance of a subject’s overall appearance throughout the long-horizon generation process.

\subsection{Environment Consistency}
Environment consistency evaluates the model's ability to maintain the stability of the scene's spatial structure and visual rendering characteristics throughout the generation process. We decompose it into spatial consistency and rendering consistency.

\subsubsection{Spatial Consistency}
Spatial consistency focuses on the consistency of the 3D spatial structure of the scene and the geometric relationship of camera motion.

\textbf{Epipolar Geometry.}
To rigorously evaluate the relative geometric constraints between views, we first utilize DA3~\cite{lin2025da3} to estimate the camera poses for all frames across the generated video. 
Instead of relying on temporally consecutive frames, we establish a robust evaluation set by sampling multiple frame pairs that exhibit similar estimated camera poses but are strictly not temporally adjacent (i.e., non-consecutive) in the video sequence. 
Let this set of sampled non-adjacent frame pairs be denoted as $\mathcal{S}$. For each frame pair in $\mathcal{S}$, we identify $N$ corresponding point pairs $\{(p_i, p_i')\}$. The Epipolar Geometry Error $E_{\text{Epipolar}}$ for a single pair is defined as:
\begin{equation}
    E_{\text{Epipolar}} = \frac{1}{2N} \sum_{i=1}^{N} \left( d(p_i, F^T p_i') + d(p_i', F p_i) \right),
\end{equation}
where $F$ is the fundamental matrix derived from the DA3-estimated poses, and $d(p, l)$ represents the pixel distance from a point $p$ to its corresponding epipolar line $l$. 
The overall Epipolar Geometry Consistency Score for the video is then computed as $\exp{(-E_{\text{Epipolar}})}$.

\textbf{Reprojection Consistency.}
Complementary to the relative epipolar constraints, Reprojection Consistency assesses the absolute fidelity of the 3D-to-2D projection. 
Leveraging the same set of non-adjacent frame pairs $\mathcal{S}$, we triangulate $N$ 3D spatial points $\{X_i\}$ based on their 2D matches. The Reprojection Error $E_{\text{Reproject}}$ quantifies the deviation of these 3D points from their 2D image observations $\{p_i\}$ when projected back onto the image plane. For a given frame, it is formulated as:
\begin{equation}
    E_{\text{Reproject}} = \frac{1}{N} \sum_{i=1}^{N} \| p_i - K[R|t]X_i \|_2, 
\end{equation}
where $K$ denotes the camera intrinsic matrix, and $[R|t]$ represents the extrinsic camera parameters explicitly provided by the DA3 estimation. The final Reprojection Consistency Score is defined as $\exp{(-E_{\text{Reproject}})}$.

\subsubsection{Rendering Consistency}
Rendering consistency focuses on the stability of the visual rendering characteristics of the scene, independent of its geometric structure.

\textbf{Lighting Consistency.}
To quantitatively evaluate the temporal stability of global brightness, color temperature, color distribution in generated videos, we introduce the Lighting Consistency Metric. This metric is particularly crucial for evaluating videos featuring \textit{departure-return} trajectories, as it assesses whether the generation model can faithfully remember and reconstruct the original lighting conditions when revisiting a scene.
For a generated video $V$, we establish a set of corresponding segment pairs $\mathcal{P} = \{(i, j)\}$, where $s_i$ denotes a temporal segment from the forward pass and $s_j$ represents its spatially matching counterpart from the return pass. 
The overall lighting deviation $D_{\text{light}}(V)$ is formulated as a weighted combination of illumination discrepancy and color shift:
\begin{equation}
    D_{\text{light}}(V) = \frac{1}{|\mathcal{P}|} \sum_{(i,j) \in \mathcal{P}} \Big[ \alpha |A_i - A_j| + (1 - \alpha) \|c_i - c_j\|_2 \Big]
\end{equation}
where the first term evaluates the preservation of structural illumination using the $L_1$ norm, and the second term measures the stability of the color distribution using the $L_2$ norm. $\alpha$ serves as a balancing hyperparameter.
To extract these components, we map the video frames into the CIELAB color space. The illumination map $A_i$ for a given segment $s_i$ is derived by 
calculating the spatial mean $\mu(\cdot)$ of the Lightness channel ($L$) to filter out high-frequency textures, followed by temporal averaging over all frames $t$ within the segment:
\begin{equation}
    A_i = \frac{1}{|s_i|} \sum_{t \in s_i} \mu(L_t)
\end{equation}
Similarly, the color shift vector $c_i$ for the segment is computed by calculating the spatial mean $\mu(\cdot)$ of the $a$ and $b$ color channels:
\begin{equation}
    c_i = \frac{1}{|s_i|} \sum_{t \in s_i} [\mu(a_t), \mu(b_t)]
\end{equation}
Finally, to convert the absolute deviation distance into a normalized consistency score that is bounded and easily interpretable, we apply an exponential decay function. The final Lighting Consistency Score is defined as:
\begin{equation}
    \text{Lighting Consistency Score} = \exp \left( - \frac{D_{\text{light}}(V)}{\tau} \right)
\end{equation}
where $\tau$ is a temperature parameter that controls the sensitivity of the metric to lighting variations. A higher $M_{\text{light}}$ score indicates better lighting and color stability throughout the generation process.

\textbf{Style Consistency.}
Style consistency evaluates whether the overall visual style of the generated video remains consistent with the historical context. We use \textit{Gram Matrix Distance} as the evaluation metric, which is widely used in style transfer tasks to measure style similarity:
\begin{equation}
\text{Style Consistency Score} = 1 - \frac{1}{T-1} \sum_{t=1}^{T-1} \| G_t - G_{t-1} \|_F
\end{equation}
where $G_t$ denotes the Gram matrix of the feature maps extracted from a pre-trained VGG network for frame $t$, and $\|\cdot\|_F$ denotes the Frobenius norm.

\subsection{Causal Consistency}
Causal consistency evaluates the model's ability to maintain the logical causal relationships between prior events and future scene dynamics, ensuring that generated content follows the physical and semantic rules established in the preceding context. We decompose it into self-evolution consistency and interaction consistency.

\subsubsection{Self-Evolution}
In the context of generative world models, self-evolution refers to the intrinsic capability of a model to autonomously simulate and progress physical dynamics, object interactions, or state transitions over time, guided solely by initial textual instructions. 
A robust world model should not only generate high-fidelity static appearances but also naturally unfold temporal events with causal consistency. To comprehensively evaluate this capability, we conceptualize self-evolution through a decoupled assessment framework comprising two fundamental perspectives: first, whether the specified physical evolution is successfully initiated (i.e., the occurrence or triggering of the event), and second, whether the ensuing dynamic progression adheres to real-world physical laws and causal logic (i.e., the correctness of the evolution). Building upon this intuition, we detail the corresponding metrics below.

\textbf{State Evolution.}
To evaluate whether a video generation model can successfully initiate a specified physical dynamic process, especially focusing on the state evolution within temporarily unseen regions during a departure-return camera trajectory, we introduce the State Evolution Score. 
Given a generative world model and a textual prompt describing a specific physical event that is expected to progress while the target is out of the camera's field of view (FoV). 
We employ a Vision-Language Model (VLM), to serve as an automated evaluator. 
The VLM maps the video-text pair to the State Evolution Score, which quantifies the degree to which the described out-of-view physical process is actually actuated upon the subject's re-entry into the frame. 
A minimal score of $1$ indicates that the event is completely un-triggered (e.g., the object or scene returns in its exact original state without the expected evolution), while a maximal score of $5$ signifies a fully realized actuation of the dynamic process during the off-screen period. Consequently, we normalize its original score into a continuous gating weight $g\in [0, 1]$.

\textbf{Evolution Correctness.}
Beyond merely triggering an event, it is crucial to address the second perspective: ensuring that the subsequent dynamic evolution adheres strictly to real-world physical laws. 
The second component of the VLM evaluation, the Evolution Correctness score $s$, assesses this physical plausibility. 
A score of $1$ denotes a complete violation of physical principles (e.g., unnatural deformations or object vanishing), whereas a $5$ represents a perfectly logical and physically correct progression. 
To formulate a comprehensive evaluation, we recognize that measuring correctness is only meaningful if the event actually occurs. 
Thus, we propose the final Evolution Correctness Score incorporating a soft-gating mechanism. Specifically, for the $i$-th generated sample, the integrated metric is then computed as the product of this gating weight and the correctness score:
\begin{equation}
    \text{Evolution Correctness Score} = g_i \cdot s_i,
\end{equation}
where $s_i\in [0, 1]$ is the normalized orginal score.
This soft-gating strategy explicitly penalizes generations that fail to trigger the required action, ensuring that a high Self-Evolution score is awarded only to videos that demonstrate both successful event actuation and physically sound progression.

\subsubsection{Interaction}
Interaction evaluates the model's ability to correctly respond to external instructions and maintain consistency during interactive processes.

\textbf{Text-conditioned Instruction.}
To quantitatively assess the temporal semantic alignment between the generated video content and the given textual instructions, we introduce the Text Interaction Metric. 
This metric evaluates the model's capability to faithfully generate and sustain visual content according to sequential prompts.
Specifically, each generated video is temporally partitioned into $M$ sequential segments corresponding to $M$ text prompts during video contiunous generation. 
To capture the temporal dynamics comprehensively, we uniformly sample $N$ frames from each segment.
We leverage the OpenCLIP vision-language model~\cite{ilharco2021openclip} to map both the visual frames and the textual prompts into a shared latent semantic space. 
For the $k$-th video segment, let $f_{\text{text}_k}^{\text{clip}}$ denote the textual feature embedding of the corresponding prompt assigned to this segment, and let $f_{\text{img}_{k,i}}^{\text{clip}}$ denote the visual feature embedding of the $i$-th sampled frame within this $k$-th segment. 
To provide a comprehensive measure of the model's instruction-following capability and cross-modal consistency throughout the continuous generation process, we compute the video-level Text Interaction Score. 
This score is formulated by calculating the macro-average of the cosine similarities between the text embedding and the corresponding frame embeddings across all $M$ segments and $N$ frames, scaled by a factor of 100:
\begin{equation}
    \text{Text Interaction Score} = \frac{1}{M \cdot N} \sum_{k=1}^{M} \sum_{i=1}^{N} \cos(f_{\text{img}_{k,i}}^{\text{clip}}, f_{\text{text}_k}^{\text{clip}}).
\end{equation}

\textbf{Action-conditioned Instruction.}
To quantitatively evaluate whether action-conditioned video generation models can faithfully adhere to prescribed action movement instructions, we formulate the Action Interaction Score. 
Given a specific action control prompt, we extract the frame-wise 6 DoF (Degrees of Freedom) twist from the camera extrinsics estimated by DA3~\cite{lin2025da3}. Let the estimated 6D twist at the $i$-th frame transition be defined as:
\begin{equation}
    \xi_i = [v_i; \omega_i] \in \mathbb{R}^6
\end{equation}
where $\omega_i \in \mathbb{R}^3$ represents the relative rotation between consecutive frames parameterized as an axis-angle vector, and $v_i = t_{i+1} - t_i \in \mathbb{R}^3$ denotes the relative translation.
Based on the predefined semantic instruction, we construct an ideal ground-truth twist $\xi_{\text{gt}}$. For instance, the initial phase of a \textit{left then right} panning motion is formalized as $\xi_{\text{gt}} =[0, 0, 0; 0, +1, 0]^{\top}$. 
Considering that different video generation models adopt heterogeneous mechanisms for action-conditioning inputs, the magnitude of the predicted camera motion often exhibits significant variance. To mitigate this scale ambiguity, we discard the magnitude constraints and employ the cosine similarity to compute the frame-level alignment score:
\begin{equation}
    \text{Action Instruction Score} = \cos(\xi_i, \xi_{\text{gt}}) = \frac{\xi_i \cdot \xi_{\text{gt}}}{\|\xi_i\| \|\xi_{\text{gt}}\|}.
\end{equation}






















\section{Experiments}

\subsection{Experimental Settings}

To comprehensively evaluate the capabilities of different generation paradigms on MBench, we select two primary categories of state-of-the-art baselines: text-conditioned world models and action-conditioned world models. 
We adopt an iterative extension setting for the text-conditioned models, including Memflow~\cite{ji2025memflow}, Self Forcing~\cite{huang2025selfforcing}, Skyreels V2~\cite{chen2025skyreelsv2}, Longlive~\cite{yang2025longlive}, Longcat-Video~\cite{longcat-video}, Cosmos-Predict 2.5~\cite{cosmos-predict2025}, Causal Forcing~\cite{zhu2026causal-forcing}, and Helios~\cite{helios}.\footnote{Self Forcing and Causal Forcing are not evaluated as standalone released long-video generators. We follow the Infinity-RoPE~\cite{yesiltepe2025infinity} implementation and use their corresponding base video generators under the same continuation protocol.} 
Specifically, these models are tasked with autoregressively predicting the future content for 5 consecutive segments, with each segment consisting of approximately 5 seconds. 
For the action-conditioned world models, consisting of Matrix-Game 2.0~\cite{matrix-game-2}, HY-WorldPlay~\cite{hyworld2025}, Yume-1.5~\cite{yume-1.5}, Matrix-Game 3.0~\cite{matrix-game-3.0}, Lingbot-World~\cite{lingbot-world}, and Infinite-World~\cite{wu2026infiniteworld}, we evaluate their performance by generating 25-second action-conditioned continuation rollouts following each model's native control interface. 
For trigger-conditioned scoring, we apply Seed 2.0~\cite{bytedance2026seed2} verify whether each generated video enters the intended memory challenge, such as an entity leaving and reappearing, a viewpoint departing and returning, or a causal event continuing through occlusion or viewpoint transfer. 
This trigger is an evaluability gate rather than a judgment of whether the model has already failed.
We evaluate self-evolution metrics with Qwen3-VL-235B-A22B~\cite{Qwen3-VL}.
All metric scores are mapped and normalized to a range of $[0, 100]$, where higher scores represent superior consistency.

\subsection{MBench Evaluation}

\definecolor{lightpurple}{rgb}{0.8, 0.6, 1.0}
\begin{table}[t]
\centering
\caption{\textbf{MBench quantitative results.} We report the M-Score for each dimension, which verify whether the model has successfully executed the memory-triggering event specified in the prompt.}
\resizebox{\linewidth}{!}{
\setlength{\tabcolsep}{4pt}
\renewcommand{\arraystretch}{1.4}
\begin{tabular}{l|cccc|cccc|cccc}
\toprule
\multirow{2}{*}{Methods} & \multicolumn{2}{c}{Object} & \multicolumn{2}{c|}{Human} & \multicolumn{2}{c}{Spatial} & \multicolumn{2}{c|}{Rendering} & \multicolumn{2}{c}{Self-evolution} & \multicolumn{2}{c}{Interaction} \\ 
\cmidrule{2-13}
& Geometry & Texture & Identity & Appearance 
& Epipolar & Reprojection & Lighting & Style 
& State & Correctness & Text & Action \\ 
\midrule
\multicolumn{13}{l}{\textit{Text-conditioned continuation}} \\
\midrule
Memflow~\cite{ji2025memflow}               & 61.72 & \cellcolor{lightpurple!30}56.06 & 39.48 & 51.60 & \cellcolor{lightpurple!30}57.95 & 20.89 & 55.06 & 30.08 & 62.75 & 72.04 & \cellcolor{lightpurple!30}46.31 & - \\
Self Forcing~\cite{huang2025selfforcing}     & 34.97 & 33.02 & \cellcolor{lightpurple!70}43.92 & 54.58 & \cellcolor{lightpurple!70}67.44 & \cellcolor{lightpurple!30}55.19 & \cellcolor{lightpurple!70}66.83 & \cellcolor{lightpurple!70}30.15 & 50.19 & 66.84 & 43.91 & - \\
Skyreels V2~\cite{chen2025skyreelsv2}        & \cellcolor{lightpurple!30}70.03 & 53.70 & 24.57 & 53.76 & 49.01 & \cellcolor{lightpurple!70}56.39 & 46.44 & 20.33 & 68.35 & 79.52 & 44.68 & - \\
Longlive~\cite{yang2025longlive}             & 63.57 & 55.41 & 42.51 & \cellcolor{lightpurple!30}55.89 & 46.68 & 27.51 & \cellcolor{lightpurple!30}59.44 & 25.26 & 70.32 & 74.69 & \cellcolor{lightpurple!70}46.97 & - \\
Longcat-Video~\cite{longcat-video}           & 46.96 & 43.13 & 26.56 & 52.98 & 28.28 & 9.26  & 56.10 & 27.46 & \cellcolor{lightpurple!70}84.17 & \cellcolor{lightpurple!70}87.83 & 46.25 & - \\
Cosmos-Predict 2.5~\cite{cosmos-predict2025} & 51.90 & 47.31 & 16.95 & 45.42 & 9.73  & 14.68 & 55.95 & 22.66 & \cellcolor{lightpurple!30}83.67 & \cellcolor{lightpurple!30}80.81 & 45.08 & - \\
Causal Forcing~\cite{zhu2026causal-forcing}  & 62.23 & 53.36 & \cellcolor{lightpurple!30}42.53 & \cellcolor{lightpurple!70}64.37 & 18.10 & 2.88  & 57.44 & \cellcolor{lightpurple!30}27.48 & 64.79 & 73.10 & 44.90 & - \\
Helios~\cite{helios}                         & \cellcolor{lightpurple!70}79.43 & \cellcolor{lightpurple!70}63.70 & 31.33 & 41.64 & 24.79 & 32.46 & 41.79 & 25.26 & 58.27 & 75.08 & 43.17 & - \\

\midrule
\multicolumn{13}{l}{\textit{Action-conditioned rollouts}} \\
\midrule
Matrix-Game 2.0~\cite{matrix-game-2}      & 14.62 & 28.99 & 1.22  & 0.94  & 14.78 & 3.08  & 38.79 & 73.78 & 10.00 & 26.40 & - & 47.86 \\
Matrix-Game 3.0~\cite{matrix-game-3.0}    & 44.15 & 58.22 & \cellcolor{lightpurple!30}42.38 & \cellcolor{lightpurple!30}47.91 & 61.99 & 32.86 & 62.06 & 95.17 & 37.50 & 48.80 & - & 81.93 \\
HY-WorldPlay~\cite{hyworld2025}           & \cellcolor{lightpurple!30}47.12 & \cellcolor{lightpurple!70}68.54 & \cellcolor{lightpurple!70}52.46 & \cellcolor{lightpurple!70}66.58 & \cellcolor{lightpurple!70}83.86 & \cellcolor{lightpurple!70}68.17 & \cellcolor{lightpurple!70}82.67 & \cellcolor{lightpurple!70}98.23 & 49.50 & 62.40 & - & \cellcolor{lightpurple!30}85.69 \\
Yume-1.5~\cite{yume-1.5}                  & \cellcolor{lightpurple!70}60.96 & 49.99 & 17.41 & 40.57 & 51.86 & 24.55 & 51.21 & 92.05 & \cellcolor{lightpurple!70}97.90 & \cellcolor{lightpurple!70}95.00 & - & 62.20 \\
Lingbot-World~\cite{lingbot-world}        & 33.20 & 44.54 & 11.57 & 33.53 & 22.12 & 7.57  & 40.06 & 85.87 & \cellcolor{lightpurple!30}96.00 & \cellcolor{lightpurple!30}89.40 & - & 63.32 \\
Infinite-World~\cite{wu2026infiniteworld} & 35.70 & \cellcolor{lightpurple!30}61.88 & 23.08 & 46.85 & \cellcolor{lightpurple!30}74.04 & \cellcolor{lightpurple!30}61.51 & \cellcolor{lightpurple!30}62.63 & \cellcolor{lightpurple!30}96.87 & 48.00 & 78.40 & - & \cellcolor{lightpurple!70}86.37 \\

\bottomrule
\end{tabular}
}
\label{tab:mbench}
\end{table}

Table~\ref{tab:mbench} reports the quantitative results on the current MBench evaluation suite. Overall, the results show that memory remains a substantial bottleneck for both text-conditioned video continuation models and action-conditioned world models. Models that produce visually plausible videos often fail to maintain a persistent world state after the target entity or camera view leaves the visible region. The failure is especially clear in spatial and causal metrics, where a model must recover a previously observed viewpoint or continue an off-screen physical process rather than merely preserve local frame quality.

\subsubsection{Text-conditioned video continuation models}

For text-conditioned continuation models, Self Forcing\cite{huang2025selfforcing} and LongLive\cite{yang2025longlive} are the most competitive baselines across several memory dimensions. Helios\cite{helios} performs best on object geometry and object texture, while Self Forcing\cite{huang2025selfforcing} excels on human identity and most environment metrics, suggesting that training with self-generated histories improves robustness to long-horizon drift. Longlive\cite{yang2025longlive} achieves the best prompt-interaction score, and Causal Forcing\cite{zhu2026causal-forcing} achieves the strongest human appearance consistency, reflecting its advantage in maintaining semantic content over multiple generated segments. MemFlow~\cite{ji2025memflow} is also competitive on object texture and spatial epipolar consistency, which is consistent with the motivation of explicitly retrieving or reusing historical visual information.

Despite these relative differences, the absolute style scores remain low for all continuation models. Even the strongest models show much weaker scores on style consistency than on epipolar or lighting. This indicates that current long video generators can often keep a stable visual tone while still forgetting the underlying 3D layout. The causal prompt-interaction scores are relatively close across models, with LongLive\cite{yang2025longlive}, Cosmos-Predict 2.5\cite{cosmos-predict2025}, MemFlow\cite{ji2025memflow}, and Longcat-Video\cite{longcat-video} clustered near the top. However, such prompt alignment does not imply physical memory: causal scores still expose large gaps in whether models continue hidden events with plausible state evolution. These results support the central design of MBench: long-term memory must be evaluated through entity, environment, and causal consistency together, rather than through visual quality or instruction following alone.

\subsubsection{Action-conditioned video world models}

Among action-conditioned world models, HY-WorldPlay\cite{hyworld2025} obtains the strongest overall performance on entity and rendering consistency. It achieves the highest human identity and appearance consistency scores, and also leads on object texture, lighting, and style consistency. This suggests that its rollout mechanism is comparatively effective at preserving both foreground entities and global rendering statistics when the camera exits and later returns to a scene. Matrix-Game 3.0\cite{matrix-game-3.0} also performs competitively on rendering style, closely following HY-WorldPlay and Infinite-World.

The spatial metrics expose a sharper gap. HY-WorldPlay\cite{hyworld2025} has the highest epipolar and reprojection scores among the evaluated action-conditioned models, followed by Infinite-World\cite{wu2026infiniteworld} and Matrix-Game 3.0\cite{matrix-game-3.0}. In contrast, Matrix-Game 2.0\cite{matrix-game-2} and Lingbot-World\cite{lingbot-world} show substantially larger geometric errors, implying that the scene does not return to the same 3D configuration even when the camera motion semantically suggests a departure-return trajectory. For interface following, Infinite-World\cite{wu2026infiniteworld} and HY-WorldPlay~\cite{hyworld2025} are strongest on action-motion alignment.

However, a deeper analysis of the causal metrics reveals a critical trade-off. Although HY-WorldPlay\cite{hyworld2025} and Infinite-World\cite{wu2026infiniteworld} excel at environment rendering and spatial consistency, their performance on self-evolution drops significantly. Both models achieve remarkably low state evolution scores compared to Yume-1.5\cite{yume-1.5} and Lingbot-World\cite{lingbot-world}. This contrast highlights a common failure mode in current action-conditioned models: the tendency to generate overly static scenes. By minimizing scene dynamics and object movements, HY-WorldPlay\cite{hyworld2025} and Infinite-World\cite{wu2026infiniteworld} successfully preserve 3D geometry and global lighting, yielding high spatial and rendering scores, but they fundamentally fail to actuate the intended physical processes or simulate meaningful state evolution. This discrepancy underscores the necessity of MBench's multi-dimensional evaluation: spatial memory and causal self-evolution must be assessed jointly to penalize models that merely "freeze" the scene to avoid temporal drift.

\begin{table}[t]
\caption{\textbf{Metric-human correlation on pairwise preferences.}
Each row is one of the twelve MBench consistency metrics
(Section~\ref{sec:mbench-evaluation}).
We aggregate three-annotator majority votes on pairwise tasks, convert
human preferences and automatic scores into model-level win ratios over eight
text-conditioned continuation models or six action-conditioned world models,
and report Spearman's $\rho$ and Kendall's $\tau$.
Entries marked ``-'' indicate no valid automatic scores for that setting.}
\label{tab:human_metric_correlation}
\small
\resizebox{\linewidth}{!}{
\setlength{\tabcolsep}{3.5pt}
\renewcommand{\arraystretch}{1.12}
\begin{tabular}{lll|cc|cc}
\toprule
 & & &
\multicolumn{2}{c|}{Text-conditioned continuation} &
\multicolumn{2}{c}{Action-conditioned rollouts} \\
\cmidrule(lr){4-5}\cmidrule(lr){6-7}
Memory consistency & Sub-dimension & Metric &
$\rho$ & $\tau$ &
$\rho$ & $\tau$ \\
\midrule
\multirow{4}{*}{Entity}
 & \multirow{2}{*}{Object}
 & Geometry Consistency & +0.97 & +0.87 & +0.97 & +0.87 \\
 & & Texture Consistency & +0.89 & +0.73 & +0.93 & +0.80 \\
\cmidrule(lr){2-7}
 & \multirow{2}{*}{Human}
 & Identity Consistency & +0.86 & +0.71 & +0.71 & +0.60 \\
 & & Appearance Consistency & +0.69 & +0.50 & +0.81 & +0.67 \\
\midrule
\multirow{4}{*}{Environment}
 & \multirow{2}{*}{Spatial}
 & Epipolar Geometry & +0.16 & +0.11 & +0.94 & +0.87 \\
 & & Reprojection Consistency & +0.45 & +0.36 & +0.94 & +0.87 \\
\cmidrule(lr){2-7}
 & \multirow{2}{*}{Rendering}
 & Lighting Consistency & -0.49 & -0.36 & +0.71 & +0.60 \\
 & & Style Consistency & +0.56 & +0.46 & +0.71 & +0.47 \\
\midrule
\multirow{4}{*}{Causal}
 & \multirow{2}{*}{Self-evolution}
 & State Progress & +0.83 & +0.71 & +0.54 & +0.47 \\
 & & Physical Plausibility & +0.81 & +0.64 & +0.43 & +0.20 \\
\cmidrule(lr){2-7}
 & \multirow{2}{*}{Interaction}
 & Text Interaction     & +0.19 & +0.11 & -- & -- \\
 & & Action Interaction & -- & -- & +0.71 & +0.60 \\
\bottomrule
\end{tabular}
}
\end{table}

\subsection{Human Alignment}
\label{sec:human_alignment}

In this section, we conduct a human study on the MBench annotation platform and compare automatic scores with human pairwise preferences over memory consistency. 
This main text focuses on the core correlation result, while more detailed trigger calibration and ablation analyses are deferred to the appendix.

\subsubsection{Human annotation protocol}
In total, the current annotation set contains 2,362 records from 22 unique annotators, including 400 binary judgments on text-conditioned videos, 1,020 pairwise judgments on text-conditioned video rollouts, and 942 pairwise judgments on action-conditioned video rollouts. 
The annotation pool covers all 14 video world models and all 12 dimensions.
The binary interface asks whether a single generated continuation video contains a memory challenge, such as transition, exit-reentry, occlusion, or departure-return camera motion that requires the model to remember temporarily invisible content. 
The pairwise interface then asks annotators to compare two model outputs on dimension-specific memory consistency questions.

\subsubsection{Metric-human correlation}
Table~\ref{tab:human_metric_correlation} reports Spearman's $\rho$ and Kendall's $\tau$ after converting both human votes and automatic scores into model-level win ratios. 
Entity consistency metrics show the strongest agreement on text-conditioned continuation, especially geometry consistency ($\rho=0.97$). 
Rendering metrics are also aligned on action-conditioned rollouts ($\rho\approx0.71$ for lighting and style), while spatial epipolar geometry is much more predictive for action-conditioned models ($\rho=0.94$) than for continuation models ($\rho=0.16$). 
Regarding the causal dimensions, self-evolution metrics show strong human correlation for text-conditioned models ($\rho>0.80$), but relatively lower alignment for action-conditioned ones ($\rho\approx0.50$), since action-driven models currently struggle to generate complex causal state changes.
Finally, interaction correlations remain setting-specific, with action interaction showing strong alignment ($\rho=0.71$) while text interaction correlates poorly ($\rho=0.19$), further validating that explicit control interfaces provide a more reliable anchor for human-metric agreement in memory evaluation.

\section{Discussion}

Our experiments suggest that current long-video generation systems are still far from reliable video world models. 
Although recent models can synthesize visually plausible and temporally extended clips, their apparent coherence often breaks when the evaluation requires a persistent world state. 
Entity appearance, background layout, and causal states may remain locally convincing, yet drift after occlusion, camera motion, or long-horizon continuation.

\textbf{Spatial understanding remains limited.}
A recurring bottleneck is spatial reasoning. On the generation side, many models struggle to follow camera-motion instructions specified in the prompt: the camera may move in the wrong direction, stop too early, or produce a visually plausible motion that does not match the requested trajectory. On the evaluation side, current vision models also have limited ability to infer camera motion from a sparse set of extracted frames. This makes spatial memory a particularly difficult axis: a model must both execute a controlled viewpoint change and recover the same 3D configuration when the view returns, while the evaluator must distinguish true geometric consistency from superficial visual similarity.

\textbf{Action-conditioned models do not yet provide stable control.}
For action-conditioned world models, the response to actions is uneven across methods. Some models can produce approximate motion under common actions, but others respond with a clear delay, ignore part of the control signal, or collapse to frequent motion patterns seen during training. In these cases, the generated video may look reasonable as a generic rollout while failing as an interactive world simulation. This suggests that action conditioning should be evaluated not only by visual quality, but also by whether the model can execute novel or composed actions faithfully over time.

\textbf{Causal memory is the central unresolved challenge.}
The causal dimension exposes an even deeper limitation. For many action-conditioned models, the generated scene remains mostly static: objects and agents may preserve their appearance, but the world itself does not evolve in a meaningful way. Such videos can obtain acceptable scores on appearance or rendering consistency while failing to simulate dynamic state transitions. This limitation becomes more severe when the relevant evolution occurs out-of-view. A reliable world model should continue hidden physical processes, remember their intermediate consequences, and reveal a plausible later state when the camera returns. Current models often lack this capability: hidden events may be ignored, reset, or replaced by visually plausible but causally unrelated content. This is why MBench separates trigger coverage from memory reliability. A model should not be rewarded for avoiding the event, and it should also not be rewarded for triggering an event without preserving its causal consequences. Future video world models therefore need stronger mechanisms for state tracking, action grounding, and latent causal simulation, rather than only better frame-level synthesis.

\section{Conclusion}

We introduced MBench, a benchmark for evaluating the memory capability of video world models across entity, environment, and causal consistency. Unlike standard video-generation evaluation, MBench explicitly tests whether a model can retain a persistent world state under long-horizon continuation, camera motion, occlusion, action conditioning, and hidden causal evolution. Through evaluations on both text-conditioned long-video generators and action-conditioned world models, we find that current systems still struggle to preserve stable entities, recover consistent spatial layouts, follow controlled actions, and simulate causal state changes that happen outside the visible field of view. These results
suggest that memory should be treated as a first-class capability for future world models, rather than as a by-product of visual fidelity or prompt alignment. We hope MBench provides a concrete diagnostic tool and a common evaluation protocol for building video world models that are not only visually plausible, but also persistent, controllable, and causally coherent.

\textbf{Acknowledgements.}
This work was supported in part by the National Natural Science Foundation of China under Grant 62576185, by the Beijing Natural Science Foundation of China under Grant L252011, and by the Young Elite Scientist Sponsorship Program by CAST under Grant YESS20240544.
We also sincerely thank Ruofeng Yang for his valuable feedback.

\bibliographystyle{plain}
\bibliography{main}

\clearpage
\appendix
\section*{Appendix}

In this supplementary material, we provide further details on the MBench benchmark construction and evaluation protocol. 
We first describe the dataset composition for both evaluation settings in Section~\ref{app:dataset}. 
Section~\ref{app:eval} elaborates on the evaluation and trigger mechanisms, including a comparison of the differences in computational strategies between Text-Conditioned Video Generation Models (MBench-T) and Action-Conditioned Video World Models (MBench-A). 
Section~\ref{app:human} details the human annotation protocol and validates the VLM trigger judge against human binary labels. 
Additional qualitative visualizations and case studies are provided in Section~\ref{app:visuals}. 

\section{Dataset and Prompt Details} \label{app:dataset}

\subsection{Data Composition}

MBench is built upon five real-world video datasets covering diverse scenarios.
Table~\ref{tab:data_composition} summarizes the number of evaluation samples per subset across the two benchmark settings. Here \textit{Human} subset and \textit{Object} subset are used to evaluate Entity-Human and Entity-Object dimensions respectively. While Environment and Causal subset is used for Causal and Environment Consistency dimensions.

\begin{table}[h]
\centering
\caption{\textbf{MBench data composition.} Number of evaluation samples per subset.}
\label{tab:data_composition}
\begin{tabular}{lrrrr|r}
\toprule
 & \textbf{Human} & \textbf{Object} & \textbf{Causal} & \textbf{Environment} & \textbf{Total} \\
\midrule
MBench-A & 120 & 100 & 100 & 227 & 547 \\
MBench-T & 120 & 100 & 200 & 73  & 493 \\
\midrule
\textbf{Total} & 240 & 200 & 300 & 300 & 1,040 \\
\bottomrule
\end{tabular}
\end{table}

The source datasets cover complementary domains to construct our evaluation subsets:
\textbf{DL3DV}~\cite{ling2024dl3dv}, \textbf{Tanks and Temples}~\cite{tanks-temples2017}, and \textbf{SpatialVID}~\cite{wang2025spatialvid} are utilized to build the environment and object subsets. 
\textbf{OpenHumanVID}~\cite{li2024openhumanvid} is employed to construct the human subset, while \textbf{Physics-aware-video}~\cite{physics-aware-video} contributes to the causal subset. 
These subsets are respectively deployed to evaluate three core memory dimensions: environment consistency, entity consistency, which comprises human and object, and causal consistency.

\section{Evaluation and Trigger Details} \label{app:eval}

For each of the 12 sub-dimensions, the core metric formula is shared between text-conditioned video continuation models and action-conditioned world models, but the preprocessing steps differ in three key aspects: (a) how frame pairs or segments are selected, (b) how the memory trigger is determined, and (c) how the target entity is detected. 

\subsection{Pair Selection Strategies}

\textbf{MBench-T: Caption-segment-based pairs.}
For text-conditioned continuation videos partitioned into $M$ caption segments, we sample frames from early segments (segment 0-1) and late segments (segment $M\!-\!2$ to $M\!-\!1$).
Candidate pairs $(i, j)$ are scored by a combination of pose-return similarity and temporal span:
the pose-return score $\exp(-\|\mathbf{t}_i - \mathbf{t}_j\| / 0.1 - \|\mathbf{R}_i - \mathbf{R}_j\|_{\text{deg}} / 10.0)$ is computed from DA3~\cite{lin2025da3} extrinsics, and pairs are ranked primarily by this score and secondarily by temporal span. The top 5 pairs are retained.

\textbf{MBench-A: Pose-based fold pairs.}
For action-conditioned videos featuring departure-return camera trajectories, we use DA3~\cite{lin2025da3} to estimate camera extrinsics for every frame.
We detect the turnaround point where the camera is farthest from its initial pose, dividing frames into outbound (departure) and inbound (return) phases.
Each outbound frame $i$ is matched to the inbound frame $j$ with the smallest combined translation and rotation distance:
\begin{equation}
    j^* = \arg\min_j\big[\,\|\mathbf{t}_i - \mathbf{t}_j\| + \|\mathbf{R}_i - \mathbf{R}_j\|_{\text{deg}}/180\,\big],
\end{equation}
where $\mathbf{t}$ denotes the camera translation vector and $\mathbf{R}$ the rotation matrix.
Up to 20 pairs with the largest temporal spans are retained for evaluation.
When no clear turnaround is detected (e.g., insufficient frames), fixed symmetric pairs are used as a fallback.

\subsection{Trigger Mechanism}

Each MBench-T metric for text-conditioned video continuation models uses a VLM to verify whether the generated video has actually entered the intended memory challenge.
The trigger judge receives 8 uniformly sampled frames from the video together with the caption segments, and returns a binary decision with a confidence score.
The trigger prompt is specific to each subset:
\begin{itemize}
    \item \textbf{Human}: ``Does the video actually create a human memory challenge, such as a person appearing, leaving/being occluded across middle segments, and later reappearing?''
    \item \textbf{Object}: ``Does the video actually create an object memory challenge for the target object, such as appearing, leaving/being occluded/changing view, and later reappearing?''
    \item \textbf{Environment}: ``Does the video actually create an environment memory challenge, such as moving away from a scene/view/layout and later revisiting the same place or layout?''
    \item \textbf{Causal}: ``Does the video actually create a causal memory challenge, such as an event or physical process that should continue while partially hidden or across later segments?''
\end{itemize}
A sample is considered evaluable for a given metric only if the trigger judge returns trigger.
Models with low trigger coverage are penalized through the M-Score harmonic mean.

For MBench-A, we omit the VLM-based semantic trigger. Unlike text-conditioned generation where memory challenges (e.g., leaving and re-entering) can be easily ignored by the model due to the unconstrained nature of abstract prompts, action-conditioned models are mainly driven by explicit spatial trajectories. These predetermined camera or action movements deterministically enforce the physical occlusion and reappearance of scenes and objects, inherently activating the memory challenge without the need for semantic verification.

\subsection{Entity Detection Details}

\textbf{MBench-T: Detection-based pipeline.}
Each object evaluation item is equipped with a semantic profile comprising a descriptive name and core visual attributes.
Grounding DINO detects the target object in the initial frame using a text query formed by concatenating these attributes and the name.
The resulting bounding box serves as a point prompt for SAM~2 to generate an initial mask, which is subsequently propagated across the video via SAM~2 tracking.
For human entities, the detection process follows the same ArcFace pipeline as MBench-A, but utilizes per-caption-segment frame sampling rather than uniform sampling.

\textbf{MBench-A: Anchor-based pipeline.}
Each evaluation item provides a ground-truth spatial anchor, which is parameterized by its center coordinates, width, and height, to explicitly specify the target object's location in the first frame.
SAM~2~\cite{ravi2024sam2} utilizes this bounding box anchor to generate a precise initial mask, which is then propagated across all sampled frames via video tracking.
For human entities, InsightFace (Buffalo\_L) is employed to detect all faces within 32 uniformly sampled frames.
The facial tracks are then constructed through greedy bipartite matching (with a distance threshold of 0.15) using rolling-average ArcFace~\cite{deng2019arcface} embeddings.

\section{Human Annotation Details} \label{app:human} 

\subsection{Annotation Protocol}

Human annotations were collected through a web-based platform featuring two specialized interfaces:

\textbf{Trigger Annotation Interface.}
Annotators view a single generated video continuation and answer whether it visibly triggers a specific memory challenge.
This interface is primarily designed as a user study to collect human ground-truth labels for validating the effectiveness and reliability of our VLM-based trigger judge.

\textbf{Metric Preference Interface.}
Annotators view two model outputs side by side (generated from the same input prompt or action sequence) and select which one better maintains a specific memory consistency property, or declare a tie.
Dimension-specific questions are presented for each of the evaluated sub-dimensions.

\begin{figure}[htbp] 
    \centering
    \begin{subfigure}[b]{0.54\textwidth}
        \centering
        \includegraphics[width=\textwidth]{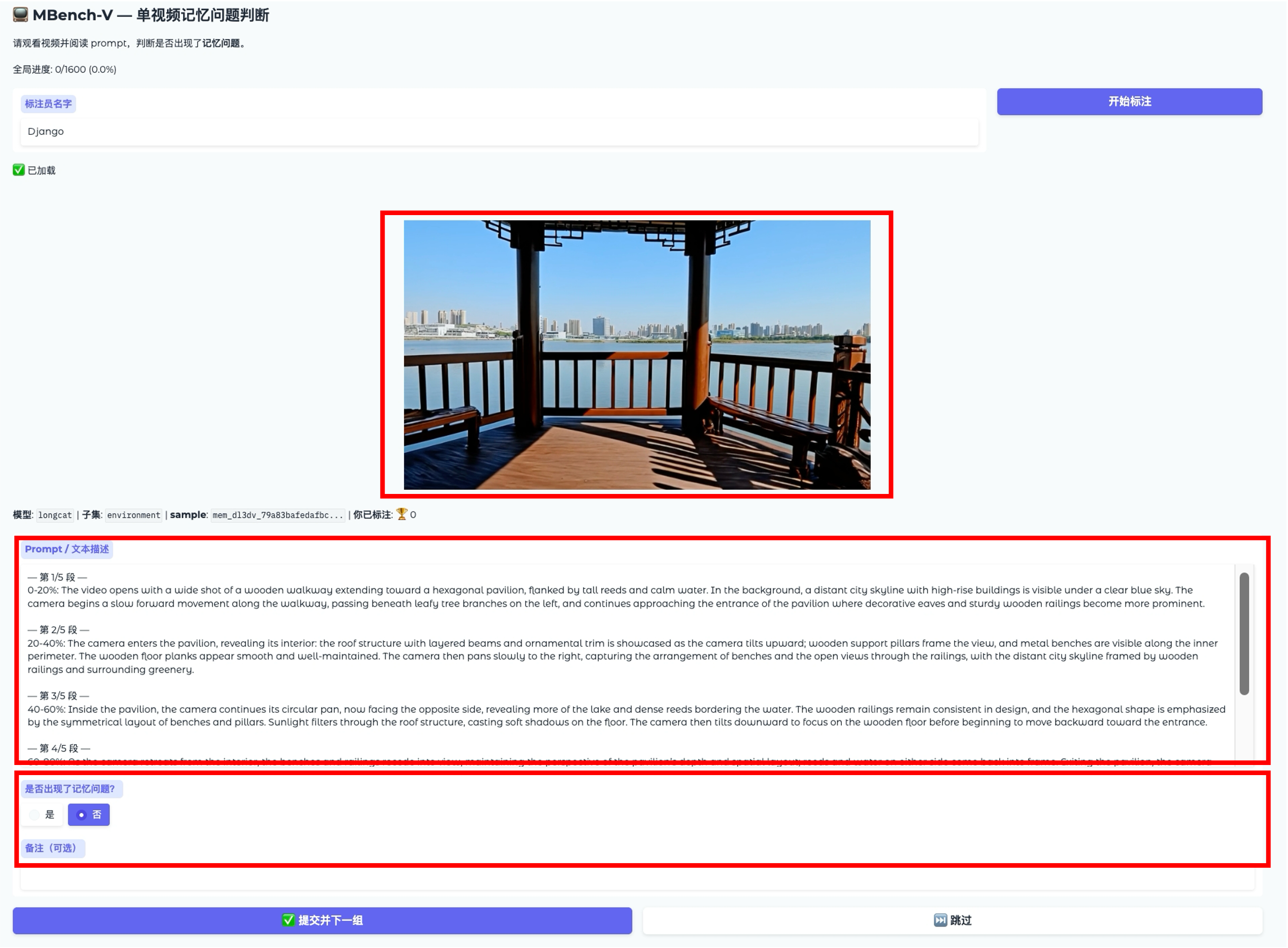}
        \caption{\textbf{Trigger Annotation Interface}}
        \label{fig:trigger_annotation}
    \end{subfigure}
    \hfill 
    \begin{subfigure}[b]{0.44\textwidth}
        \centering
        \includegraphics[width=\textwidth]{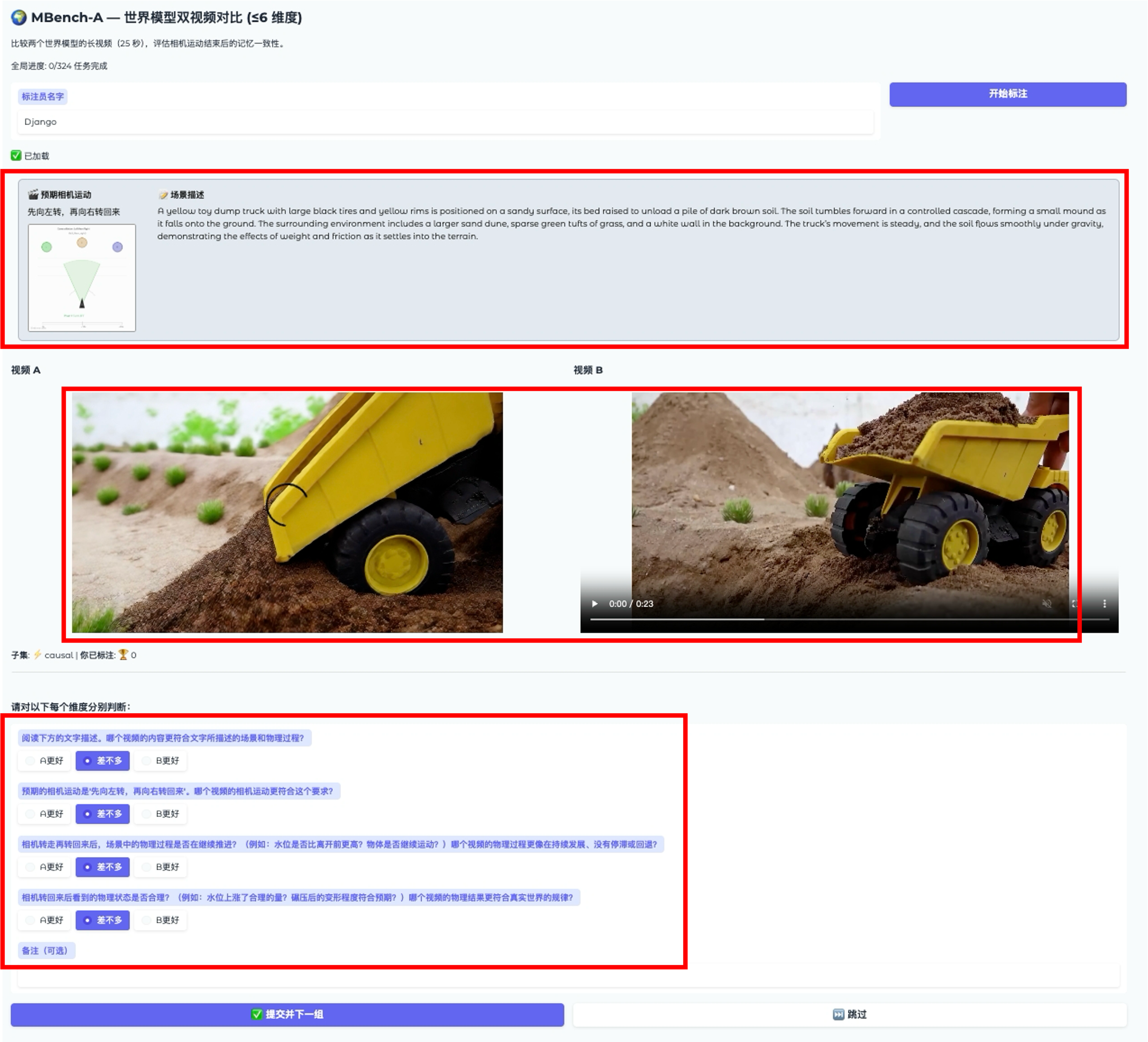}
        \caption{\textbf{Metric Preference Interface}}
        \label{fig:metric_annotation}
    \end{subfigure}
    \caption{\textbf{Human Annotation Platform.} \textbf{(a) Trigger Annotation}: The top section displays the generated video, the middle shows the 5-segment continuation captions used for generation, and the bottom provides options to determine whether the intended memory challenge is successfully triggered. \textbf{(b) Metric Preference}: The top section presents contextual information (including captions, potential camera trajectories, and object masks/descriptions). The middle displays a side-by-side comparison of two generated videos, while the bottom lists specific questions corresponding to the memory dimensions evaluated in that particular subset.}
    \label{fig:annotation_platform}
\end{figure}

\subsection{Annotation Statistics}

Table~\ref{tab:annotation_stats} summarizes the annotation pool used in this work. The pairwise comparisons evaluate the generated models across 12 fine-grained memory dimensions. To ensure the reliability of the collected data, annotator agreement is determined via a majority vote mechanism. Each video pair is independently evaluated by three annotators, and the majority consensus serves as the ground-truth human preference. Instances resulting in a complete tie (e.g., all three annotators provide divergent responses) are explicitly excluded from the final win-ratio calculations.

\begin{table}[h]
\centering
\caption{\textbf{Human annotation statistics.}}
\label{tab:annotation_stats}
\begin{tabular}{lrrr}
\toprule
 & \textbf{MBench-A Preference} & \textbf{MBench-T Preference} & \textbf{Trigger Validation} \\
\midrule
Total records        & 942  & 1,020 & 400 \\
Unique annotators    & 17   & 11    & 4 \\
Models covered       & 6    & 8     & 8 \\
\bottomrule
\end{tabular}
\end{table}

\subsection{Validation of VLM-based Triggers}

To validate the reliability of the VLM-based trigger judge, we conducted a user study collecting 400 human annotations across all four subsets (causal, object, environment, human).
Annotators were asked whether a generated video visibly enters the intended memory challenge.
Table~\ref{tab:trigger_validation} reports the accuracy, precision, recall, and F1-score of the VLM trigger decisions against the human majority vote.

\begin{table}[h]
\centering
\caption{\textbf{Validation of VLM trigger against human labels.} In our evaluation settings, the VLM achieves high recall ($\ge 0.73$) across all subsets, demonstrating its effectiveness in ensuring that genuine memory challenges are rarely missed. A false positive merely admits a sample into the downstream metric evaluation, whereas a false negative excludes genuinely informative data. With an overall recall of 0.90, the VLM serves as an effective gatekeeper for our benchmark.}
\label{tab:trigger_validation}
\begin{tabular}{lrrrr}
\toprule
\textbf{Subset} & \textbf{Accuracy} & \textbf{Precision} & \textbf{Recall} & \textbf{F1} \\
\midrule
Causal      & 0.75 & 0.75 & 1.00 & 0.86 \\
Object      & 0.86 & 0.83 & 1.00 & 0.91 \\
Environment & 0.45 & 0.44 & 0.89 & 0.59 \\
Human       & 0.70 & 0.73 & 0.73 & 0.73 \\
\midrule
\textbf{Overall} & \textbf{0.66} & \textbf{0.65} & \textbf{0.90} & \textbf{0.76} \\
\bottomrule
\end{tabular}
\end{table}

\begin{figure}[t]
    \centering
    \begin{subfigure}{1.0\linewidth}
        \centering
        \includegraphics[width=\linewidth]{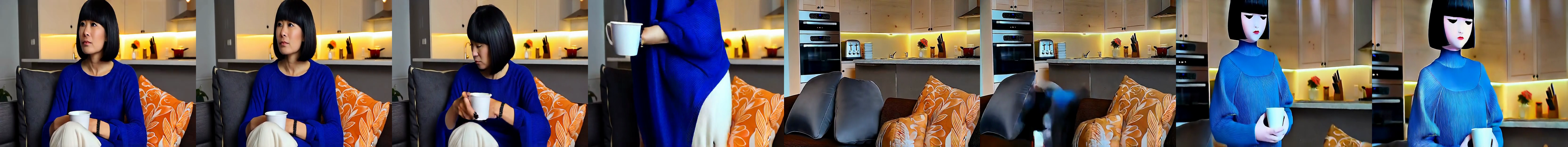}
        \caption{\textbf{Triggered Example.} Appearance Score: 10.37. Although the generated video exhibits relatively weaker visual quality, it successfully follows the prompt and produces the required disappearance--reappearance event of the target character.}
        \label{fig:triggered_case}
    \end{subfigure}
    \vspace{0.5em}
    \begin{subfigure}{1.0\linewidth}
        \centering
        \includegraphics[width=\linewidth]{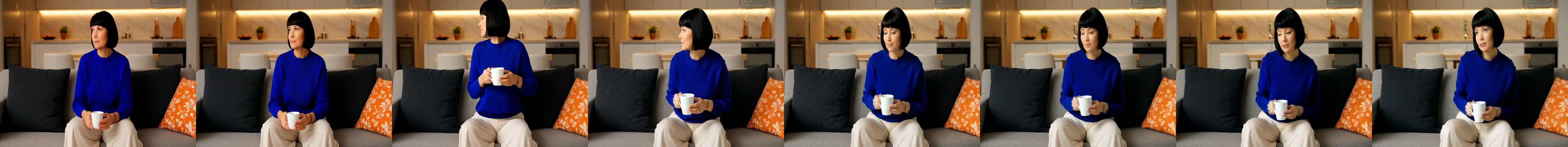}
        \caption{\textbf{Non-triggered Example.} The generated character appearance is better preserved, but the required disappearance--reappearance event never occurs. The sample is considered non-triggered; its consistency scores are excluded from final score aggregation.}
        \label{fig:nontriggered_case}
    \end{subfigure}
    \caption{\textbf{Trigger mechanism for text-conditioned generation.}
    A triggered sample contributes its consistency scores to the reliability calculation; a non-triggered sample is excluded and contributes only to the trigger coverage penalty. The full prompt used for these examples is documented in the figure source. Both examples are generated using the same five-stage prompt sequence: \textit{(1) A woman in her late 30s or early 40s stands closely embraced with a man in a softly lit office. She has shoulder-length dark brown hair, a few strands framing her face, wearing a white blouse with thin black vertical stripes under a dark blazer. Her left wrist displays a sleek black watch with a metallic band, and she holds the blazer tightly around him. (2) The woman begins to step away, moving steadily toward the right side of the frame. Her body turns slightly, revealing more of her profile as she glides forward. The blazer drapes over her arm, and the watch catches the light as her hand lifts gently. Only half of her remains visible as she edges out. (3) The office space is now empty—bookshelves lined with books and binders stand quietly, a framed abstract painting hangs on the wall, and soft natural light filters through unseen windows. The air feels still, the embrace gone, leaving only the calm, untouched environment behind. (4) From the left edge of the frame, the woman re-enters slowly, her silhouette emerging into view. Her dark blazer appears first, followed by the curve of her shoulder and the flash of her watch. Her hair falls gently across her face as she steps forward, partially visible again. (5) The woman is fully back in frame, standing centered once more. Her shoulder-length dark brown hair frames her face, the white striped blouse beneath the dark blazer, and the sleek black watch gleams on her left wrist. She is exactly as before—still, poised, and complete.}}
    \label{fig:trigger_mechanism}
\end{figure}

\section{More Visualizations and Case Studies} \label{app:visuals}

In this section, we provide qualitative visualizations to illustrate the effectiveness of the trigger mechanism and the discriminative power of MBench metrics across all three memory dimensions.

\subsection{Trigger Mechanism Visualization}

Figure~\ref{fig:trigger_mechanism} illustrates the trigger mechanism in action with two contrastive examples, both generated from the same five-stage caption prompt sequence describing a woman stepping away from and later returning to an office scene.
The first example (\ref{fig:triggered_case}) successfully triggers the memory challenge: although the generated video exhibits relatively weak appearance preservation, it faithfully follows the prompt and produces the required disappearance--reappearance event.
The second example (\ref{fig:nontriggered_case}) fails to trigger: the generated character maintains better visual quality, but the required exit--reentry event never occurs. As a result, this sample is excluded from consistency evaluation and contributes to the trigger coverage penalty in the M-Score.

\subsection{Metric Scores Visualization}

To demonstrate the discriminative power of our proposed metrics, we provide qualitative visual comparisons of high-scoring (successful) and low-scoring (failed) generated videos across the three core memory dimensions. For a fair comparison, each good--bad pair is generated conditioned on the identical input prompt or action sequence.

\begin{figure}[htbp]
    \centering
    \begin{subfigure}{\linewidth}
        \centering
        \includegraphics[width=\linewidth]{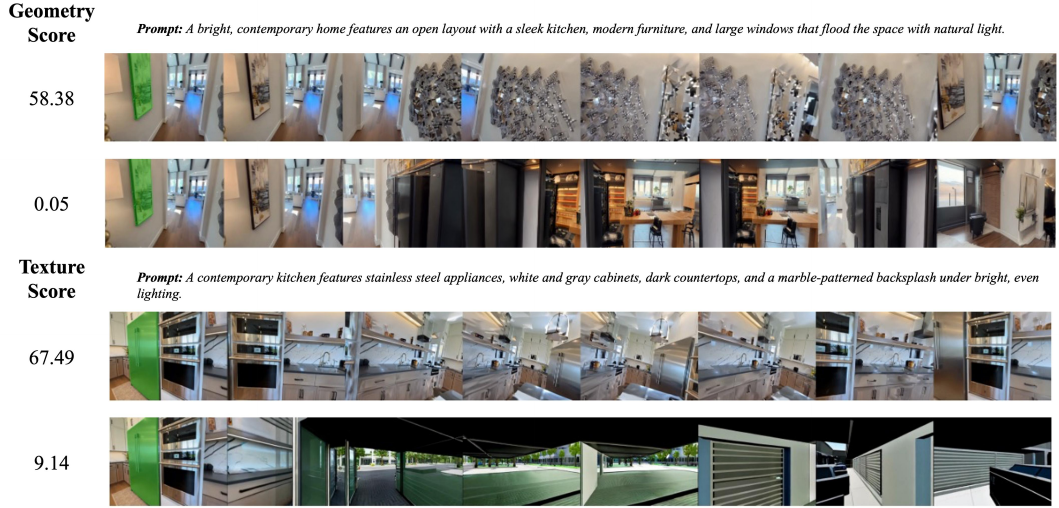}
        \caption{\textbf{Object Consistency}}
    \end{subfigure}
    \\
    \vspace{0.4cm}
    \begin{subfigure}{\linewidth}
        \centering
        \includegraphics[width=\linewidth]{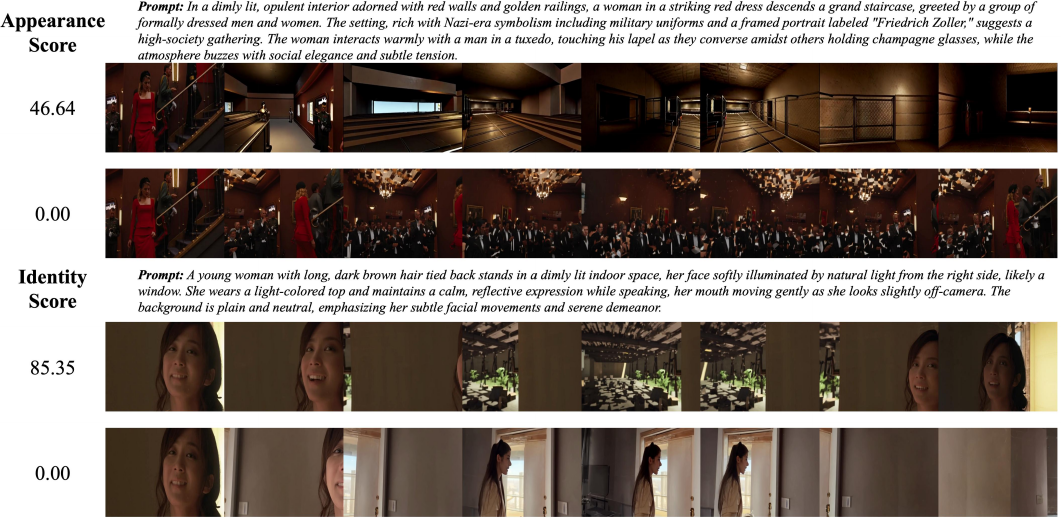}
        \caption{\textbf{Human Consistency}}
    \end{subfigure}
    \caption{\textbf{Entity consistency visualization.} Qualitative comparison of human and object consistency under identical condition. \textbf{(a) Human Consistency}: Evaluated via Appearance and Identity scores. Low-scoring examples exhibit severe garment changes or facial distortion. \textbf{(b) Object Consistency}: Evaluated via Geometry and Texture scores.}
    \label{fig:entity_viz}
\end{figure}

\textbf{Entity Consistency.} 
Figure~\ref{fig:entity_viz} illustrates the evaluation of human and object consistency. The metrics effectively distinguish between models that preserve fine-grained facial and geometric details across the departure--return trajectory, and those that suffer from severe identity drift or object deformation upon re-entry.

\begin{figure}[htbp]
    \centering
    \begin{subfigure}{\linewidth}
        \centering
        \includegraphics[width=\linewidth]{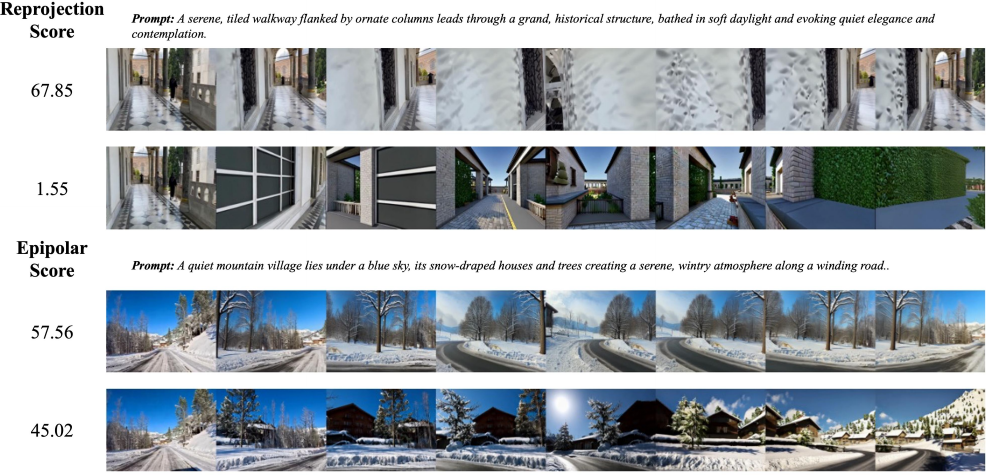}
        \caption{\textbf{Geometry Consistency}}
    \end{subfigure}
    \\
    \vspace{0.4cm}
    \begin{subfigure}{\linewidth}
        \centering
        \includegraphics[width=\linewidth]{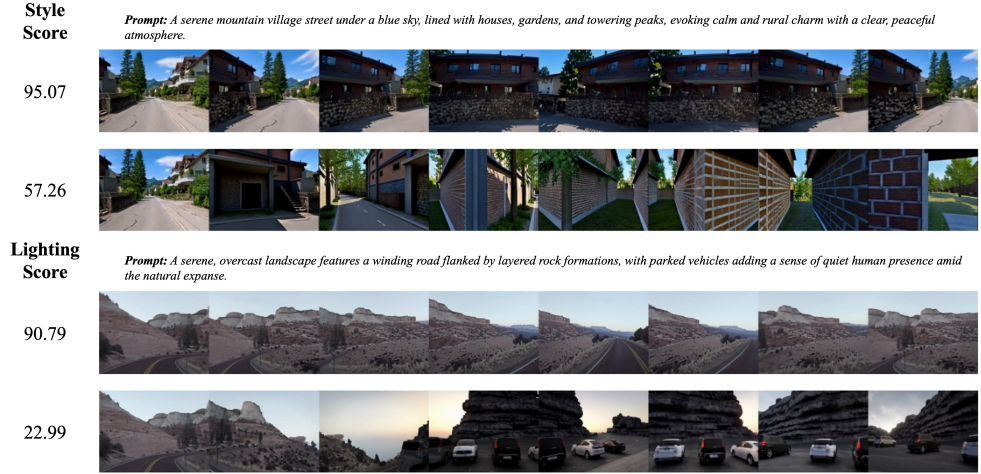}
        \caption{\textbf{Rendering Consistency}}
    \end{subfigure}
    \caption{\textbf{Environment consistency visualization.} Qualitative comparison of scene revisits. \textbf{(a) Spatial Consistency}: Measured by Reprojection and Epipolar scores. Poor models fail to maintain 3D multi-view geometry, resulting in misaligned layouts. \textbf{(b) Rendering Consistency}: Measured by Style and Lighting scores. Low-scoring examples show obvious shifts in global illumination and texture style upon returning to the initial viewport.}
    \label{fig:environment_viz}
\end{figure}

\textbf{Environment Consistency.} 
Figure~\ref{fig:environment_viz} visualizes spatial and rendering consistency during scene revisits. High scores accurately reflect mathematically preserved 3D layouts and global illumination, whereas low scores successfully capture severe geometric collapse and stylistic hallucinations.

\begin{figure}[htbp]
    \centering
    \includegraphics[width=\linewidth]{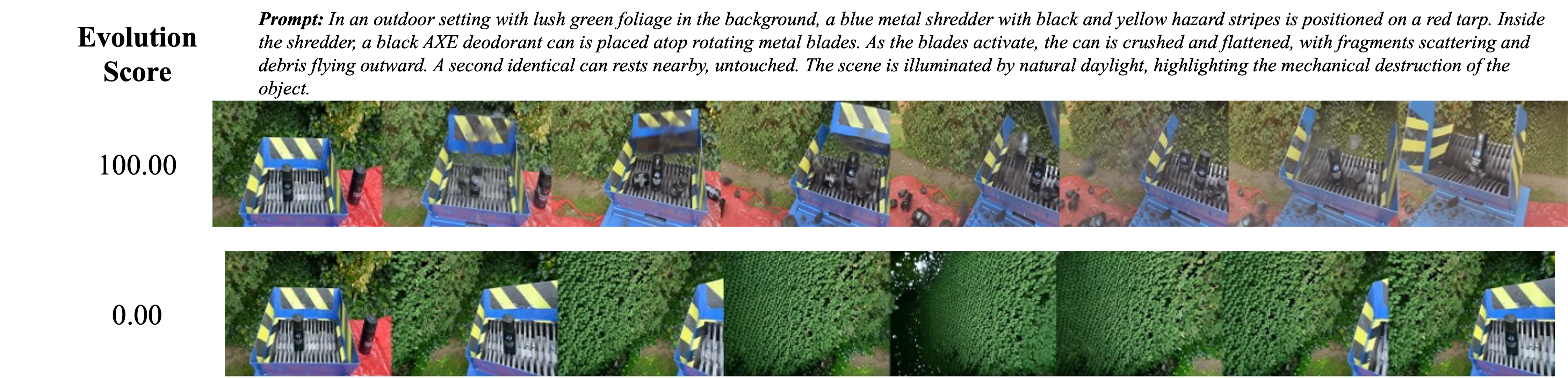}
    \caption{\textbf{Causal evolution visualization.} Qualitative comparison of state progression. The top low-scoring example (Score: 0.00) completely hallucinates the physical process, morphing the shredder into foliage. The bottom high-scoring example (Score: 100.00) correctly simulates the expected physical destruction of the deodorant can inside the shredder.}
    \label{fig:causal_evo_viz}
\end{figure}

\begin{figure}[htbp] 
    \centering
    \begin{subfigure}[b]{0.70\textwidth}
        \centering
        \includegraphics[width=\textwidth]{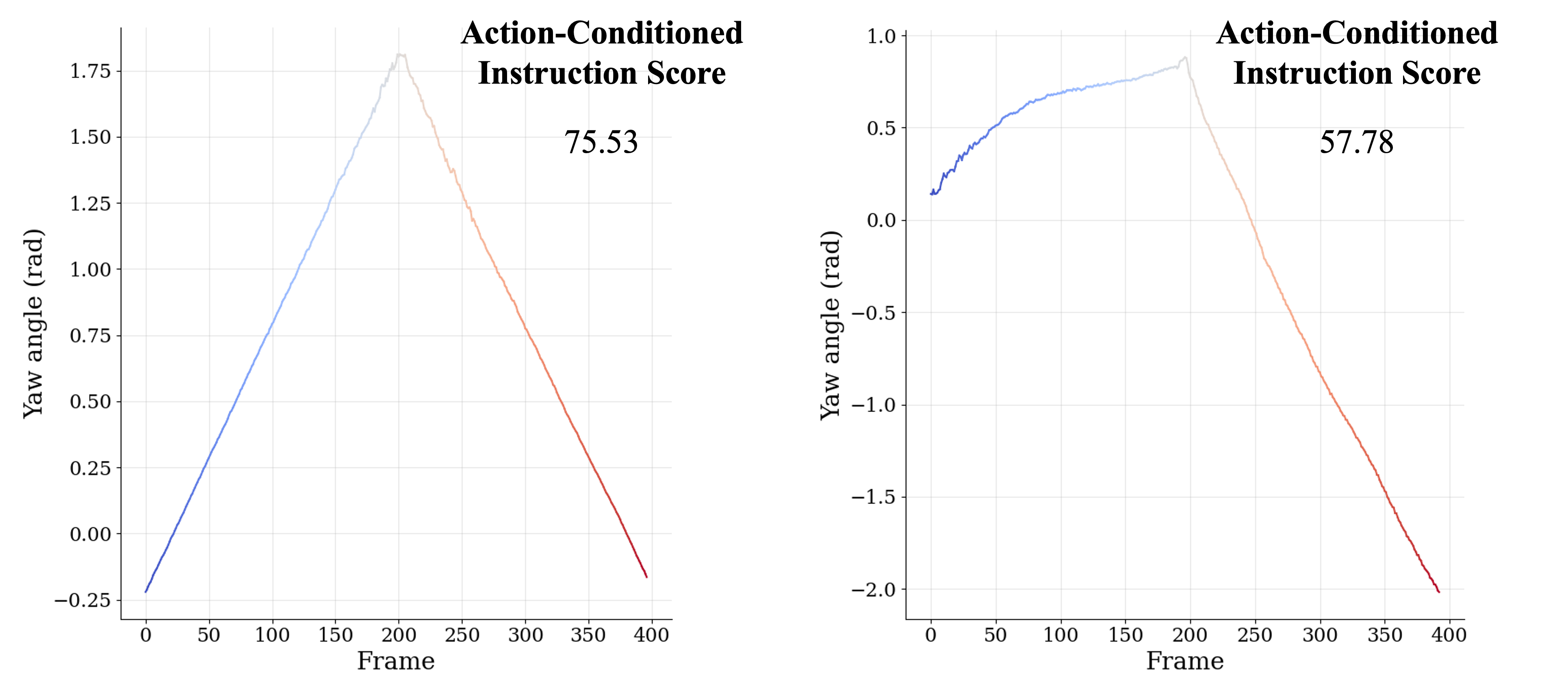} 
        \caption{\textbf{Extracted Camera Trajectories vs. Action Scores}}
        \label{fig:act_real}
    \end{subfigure}
    \hfill 
    \begin{subfigure}[b]{0.29\textwidth}
        \centering
        \includegraphics[width=\textwidth]{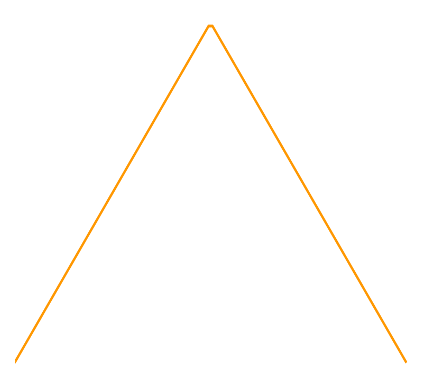}
        \vspace{0.5cm}
        \caption{\textbf{Ideal Input}}
        \label{fig:act_ideal}
    \end{subfigure}
    \caption{\textbf{Action instruction adherence visualization.} \textbf{(a)} Extracted camera trajectories (yaw angle over frames) from generated videos. The left model tightly follows the outbound-and-return instruction, yielding a high score (75.53). The right model exhibits noisy and deviated movements, resulting in a lower score (57.78). \textbf{(b)} The ideal triangular input trajectory provided as the action condition.}
    \label{fig:causal_act_viz}
\end{figure}

\begin{figure}[htbp]
    \centering
    \includegraphics[width=\linewidth]{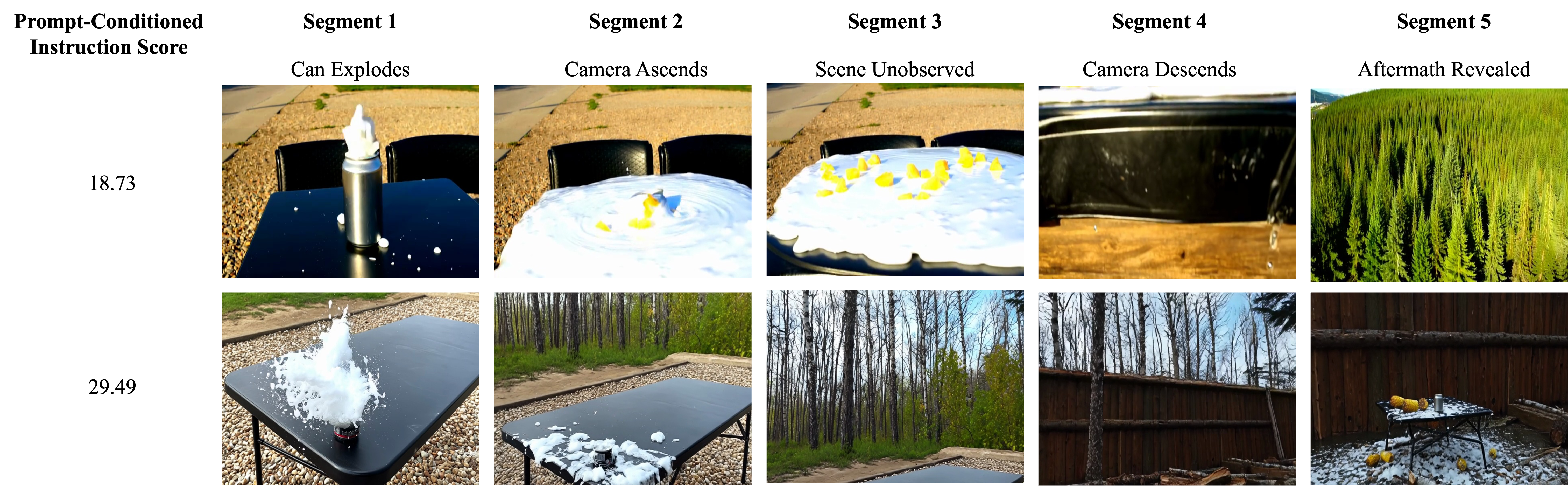}
    \caption{\textbf{Prompt instruction adherence visualization.} Qualitative comparison of text-conditioned sequential instruction following across 5 prompt segments. The top row (Score: 18.73) fails the challenge by completely hallucinating an unrelated drone-view forest in Segment 5 instead of revealing the table aftermath. The bottom row (Score: 29.49) successfully maintains scene consistency, returning to the initial black table and forest background in the final segment as instructed.}
    \label{fig:causal_prompt_viz}
\end{figure}

\textbf{Causal Consistency.} 
We visualize causal self-evolution in Figure~\ref{fig:causal_evo_viz}, highlighting the metrics' ability to heavily penalize physically implausible state transitions (e.g., hallucinatory morphing instead of physical destruction). Furthermore, Figures~\ref{fig:causal_act_viz} and~\ref{fig:causal_prompt_viz} demonstrate the evaluation of interaction adherence under action and text conditions, respectively. These metrics effectively penalize noisy motion control and semantic scene drift during long-term sequential execution.

\end{document}